\documentclass[11pt]{article}

\usepackage{epsfig}
\usepackage{amssymb}
\usepackage{natbib}
\usepackage{hyperref}
\usepackage{graphicx}
\usepackage{thumbpdf}
\usepackage{amsfonts,amstext,amsmath}
\usepackage{accents}
\usepackage{color}
\usepackage{rotating}
\usepackage[utf8]{inputenc}
\usepackage{booktabs}
\usepackage{tikz}
\usepackage{multirow}
\usepackage{float}
\usepackage{chngcntr}
\usepackage[labelfont=bf, width=15cm, font=small]{caption}
\usepackage[onehalfspacing]{setspace}
\usepackage[left=3cm, right=3cm, top=2.5cm, bottom=3cm]{geometry}
\setcitestyle{authoryear,open={(},close={)}}

\usepackage{bbm}
\usepackage{accents}

\newcommand{\1}{\mathbbm{1}}

\newcommand{\rZ}{Z}
\newcommand{\rY}{Y}

\newcommand{\rz}{z}
\newcommand{\ry}{y}
\newcommand{\rx}{\xvec}

\newcommand{\pZ}{F_\rZ}
\newcommand{\pY}{F_\rY}

\newcommand{\pN}{\Phi}
\newcommand{\pSL}{F_{\SL}}
\newcommand{\pMEV}{F_{\MEV}}
\newcommand{\pGumbel}{F_{\text{Gumbel}}}

\newcommand{\h}{h}

\newcommand{\eparm}{\vartheta}

\newcommand{\shiftparm}{\betavec}

\newcommand{\eshiftparm}{\beta}

\newcommand{\ie}{\textit{i.e.}~}
\newcommand{\eg}{\textit{e.g.}~}

\newcommand{\prob}{\mathbb{P}}
\newcommand{\Ex}{\mathbb{E}}

\usepackage{dsfont}
\newcommand{\I}{\mathds{1}}

\newcommand{\given}{\rvert}

 \DeclareMathOperator{\logit}{logit}

 \DeclareMathOperator{\iid}{i.i.d.}

 \DeclareMathOperator{\softmax}{softmax}
 \DeclareMathOperator{\odds}{odds}
 \DeclareMathOperator{\OR}{OR}
 \DeclareMathOperator{\KL}{KL}
 \DeclareMathOperator{\RPS}{RPS}
 \DeclareMathOperator{\BS}{BS}
 
 \DeclareMathOperator{\ND}{N}

 \DeclareMathOperator{\SL}{L}
 \DeclareMathOperator{\MEV}{MEV}

\def \xvec {\boldsymbol x}

 \def \calL {\mathcal L}
 
 \def \calN {\mathcal N}

\def \betavec         {\text{\boldmath$\beta$}}

\def \B {\mathsf{B}}

\def \linpred {\rx^\top\shiftparm}

\newcommand{\pkg}[1]{\textbf{#1}}
\newcommand{\proglang}[1]{\textsf{#1}}
\newcommand{\cmd}[1]{\texttt{#1()}}

\usepackage{tikz}

\usetikzlibrary{positioning, calc, shapes.geometric, shapes.multipart,
  shapes, arrows.meta, arrows, quotes,
  decorations.markings, external, trees}
\tikzstyle{line} = [draw, -latex']
\tikzstyle{Arrow} = [
        thick,
        decoration={
                markings,
                mark=at position 1 with {
                        \arrow[thick]{latex}
} },
        shorten >= 3pt, preaction = {decorate}
        ]

\begin{document}

\title{\bf 
       Deep and Interpretable Regression Models for Ordinal Outcomes}

\author{Lucas Kook$^{1,2}\footnote{Authors contributed equally}$\ , 
        Lisa Herzog$^{1,2*}$,
        Torsten Hothorn$^{1}$,
        Oliver D\"{u}rr$^{3}$,
        Beate Sick$^{1,2\footnote{Corresponding author, sick@zhaw.ch, beate.sick@uzh.ch}}$}

\date{\small
        $^1$ University of Zurich, Switzerland\\
        $^2$ Zurich University of Applied Sciences, Switzerland\\
        $^3$ Konstanz University of Applied Sciences, Germany
}

\maketitle

\begin{abstract}
Outcomes with a natural order commonly occur in prediction tasks and often the
available input data are a mixture of complex data like images and tabular predictors.
Deep Learning (DL) models are state-of-the-art for image classification tasks 
but frequently treat ordinal outcomes as unordered and lack interpretability.
In contrast, classical ordinal regression models consider the outcome's order and yield 
interpretable predictor effects but are limited to tabular data.
We present ordinal neural network transformation models ({\sc ontram}s), 
which unite DL with classical ordinal regression approaches.
{\sc ontram}s are a special case of transformation models and trade off flexibility
and interpretability by additively decomposing the transformation function into terms 
for image and tabular data using jointly trained neural networks.
The performance of the most flexible {\sc ontram} is by definition equivalent to
a standard multi-class DL model trained with cross-entropy while being faster in training when 
facing ordinal outcomes.
Lastly, we discuss how to interpret model components for both tabular and image data
on two publicly available datasets.
\end{abstract}



\section{Introduction} \label{sec:intro}

Many classification problems deal with classes that show a natural order.
This includes for example patient outcome scores in clinical studies or movie ratings
\citep{oghina2012predicting}.
These ordinal outcome variables may not only depend on interpretable tabular 
predictors like age or temperature but also on complex input data such as medical images,
networks of public transport, textual descriptions, or spectra. Depending on the
complexity of the input data and the concrete task, different analysis approaches
have established to tackle the ordinal problems.

Ordinal regression as a probabilistic approach has been studied for more than four
decades \citep{mccullagh1980regression}. 
The goal is to fit an interpretable
regression model, which estimates the conditional distribution of an ordinal outcome
variable $Y$ based on a set of tabular predictors. The ordinal outcome $Y$ can take
values in a set of ordered classes and the tabular predictors are scalar and interpretable
like age. Ordinal regression models make use of the 
information contained in the order of the outcome and provide a valid probability
distribution instead of a single point estimate for the most likely outcome. 
This is essential to reflect uncertainty in the predictions. 
Moreover, the estimated model parameters are interpretable as the effect a 
single predictor has on the outcome given the remaining predictors are held constant.
This allows experts to assess whether the model corresponds to their field
knowledge and provides the necessary trust for application in critical decision
making. 
However, there is a trade-off between interpretability and model complexity. 
The higher the complexity of a model, the harder it becomes to directly interpret 
the individual model parameters.

Deep Learning (DL) approaches have gained huge popularity over the last decade and
achieved outstanding performances on complex tasks like image classification
and natural language processing \citep{goodfellow2016deep}. The models take the
raw data as input and learn relevant features during the training procedure by
transforming the input into a latent representation, which is suitable to solve
the problem at hand. This avoids the challenging task of feature engineering, which
is necessary when working with statistical models. Yet, unlike statistical models,
most DL models have a black box character, which makes it hard to interpret individual 
model components.
In addition, ordinal outcomes in DL approaches are frequently
modeled in the same way as unordered outcomes using multi-class classification (MCC). 
That is, softmax is used as the last-layer activation and the loss 
function is the categorical cross-entropy.
In this setting, solely the probability assigned to the true class is entering 
the loss function, which ignores the outcome’s natural order.

To the best of our knowledge, there is currently no ordinal DL model, which enables 
to integrate tabular predictors and yields interpretable effect estimates for the tabular 
and the image data. 
This is a major disadvantage for example in fields like medicine which requires multiple 
data modalities for decision making but also a reliably interpretable model which quantifies
the effects of the predictors on the outcome~\citep{rudin2018stop}.

\subsection{Our Contribution} \label{sec:contrib}

In this work we introduce ordinal neural network transformation models ({\sc ontram}s),
which unite classical ordinal regression with DL approaches while
conserving the interpretability of statistical
and flexibility of DL models. We use a theoretically sound maximum-likelihood
based approach and reparametrize the categorical cross-entropy loss to incorporate
the order of the outcome. This guarantees the estimation of a valid probability distribution. 
By definition, the reparameterized negative log-likelihood (NLL) loss is able to achieve the same 
prediction performance as a standard
DL model trained with cross-entropy loss, but allows a faster training in case of an 
ordinal outcome.
The main advantage of the proposed {\sc ontram}s is that {\sc ontram}s provide 
interpretable effect estimates for the different input data, which is not possible 
with other DL models.

We view ordinal regression models from a transformation model perspective
\citep{hothorn2014conditional,sick2020deep}. This change of perspective is useful
because it allows a holistic view on regression models, which easily extends beyond
the case of ordinal outcomes.
In transformation models the problem of estimating a conditional outcome
distribution is translated into a problem of estimating the parameters of a monotonically increasing
transformation function, which transforms the potentially complex outcome distribution to 
a simpler, predefined distribution $\pZ$ of a continuous variable.

The goal of {\sc ontram}s is to estimate a flexible outcome distribution based on a set
of predictors including images and tabular data while keeping components of the model interpretable.
{\sc ontram}s are able to seamlessly integrate both types
of data with varyingly complex interactions between the two, by taking a modular
approach to model building. The data analyst can choose the scale on which to interpret
image and tabular predictor effects, such as the odds or hazard scale, by
specifying the simple distribution function $\pZ$. In addition, the data analyst has full control
over the complexity of the individual model components. The discussed {\sc ontram}s will
contain at most three (deep) neural networks for the intercepts in the transformation function, the
tabular and the image data. Together with the simple distribution function $\pZ$ the output of
these neural networks will be used to evaluate the NLL loss.
In the end, the NNs, which control the components of the model, are jointly 
fit by standard deep learning algorithms based on stochastic gradient descent. 
In this work, we feature convolutional neural networks (CNNs) for complex input data like images.
However, the high modularity of {\sc ontram}s enables many more applications such as
recurrent neural networks for text-based models.

\subsection{Organization of this Paper} \label{sec:orga}

We first describe the necessary theoretical background of multi-class classification and
ordinal regression. 
Afterwards, an outline of related work is given in Section~\ref{sec:related} to highlight 
the contributions of {\sc ontram}s to the field. 
We then provide details about {\sc ontram}s in Section~\ref{sec:ontram}. 
Subsequently, we describe the data sets, experiments, and models we use to study and benchmark 
{\sc ontram}s (Section~\ref{sec:experiments}).
We end this paper with a discussion of our results and juxtaposition of the different
approaches in light of model complexity, interpretability, and predictive performance.
We present further results in Appendix~\ref{app:learnspeed} and complement our discussion
of different loss functions and evaluation metrics in 
Appendices~\ref{app:qwk}~and~\ref{app:scoring}, respectively. Because most state-of-the-art 
approaches to ordinal outcomes are classifiers, we particularly highlight the distinction
between ordinal classification and the proposed regression approach of {\sc ontram}s in 
Appendix~\ref{app:resqwk}.

\section{Background} \label{sec:background}

First, we will describe the multi-class classification approach frequently applied
in deep learning, which will serve as a baseline model in our experiments.
We outline cumulative ordinal regression models as the statistical approach
to ordinal regression and how they can be viewed as a special case of
transformation models. In the end we summarize related work in the field of
deep ordinal regression and classification and interpretable machine learning.

\subsection{Multi-class Classification} \label{sec:mcc}

Deep learning prediction models for multi-class classification (MCC) are based on a NN whose 
output layer contains $K$ units and uses a softmax as the last-layer activation function to 
predict a probability 
for each class.
The softmax function is defined as
\begin{align}
p_k = \softmax(s_k) = \frac{\exp(s_k)}{\sum_{j=1}^K \exp(s_j)},
\end{align}
where $s_k$ is the value of unit $k$ before applying the activation function.
Typically, a MCC model is trained by minimizing the categorical cross-entropy loss.
The categorical cross-entropy is computed over $n$ samples by
\begin{align} \label{eq:cce}
\text{CE} := - \sum_{i=1}^n \sum_{k=1}^K \ry_{ki} \log \prob(\rY=\ry_{ki} \given \rx_i)
  = - \sum_{i=1}^n \log \calL_i,
\end{align}
with $\ry_{ki}$ being entry $k$ of the one-hot-encoded outcome for input $\rx_i$,
which is one for the observed class and zero otherwise. The probability for the
observed class is the likelihood contribution $\calL_i$ of the $i$th observation.
It is well known that the cross-entropy loss is equivalent to the
negative log-likelihood for multinomial outcome variables (see Figure~\ref{fig:ordinal}~A and B). 
Moreover, Eq.~\eqref{eq:cce} highlights that the likelihood is a local measure 
because solely the predicted probability for the observed class contributes to the loss.
That is, in case of ordinal outcomes, the outcome's natural order is ignored.
Note that MCC models as defined above are referred to as multinomial regression in
the statistical literature, highlighting the fact that they predict an entire probability
distribution instead of a single class (see Appendix~\ref{app:resqwk}).

\subsection{Ordinal Regression Models}

Ordinal regression aims to characterize the whole conditional distribution of an
ordinal outcome variable given its predictors.
A discussion of ordinal regression and how they can be viewed as transformation 
models is given in the following. 

\subsubsection{Cumulative Ordinal Regression Models}

The goal of the statistical approach to ordinal regression is mainly to develop
an interpretable model for the conditional distribution of the ordered outcome
variable $\rY$ with $K$ possible values $y_1 < y_2 < \dots < y_K$.
Before discussing how interpretable model components are achieved, we discuss
the question of how the order of the outcome can be taken into account.

\begin{figure}[!ht]
\center
\includegraphics[width=0.7\textwidth]{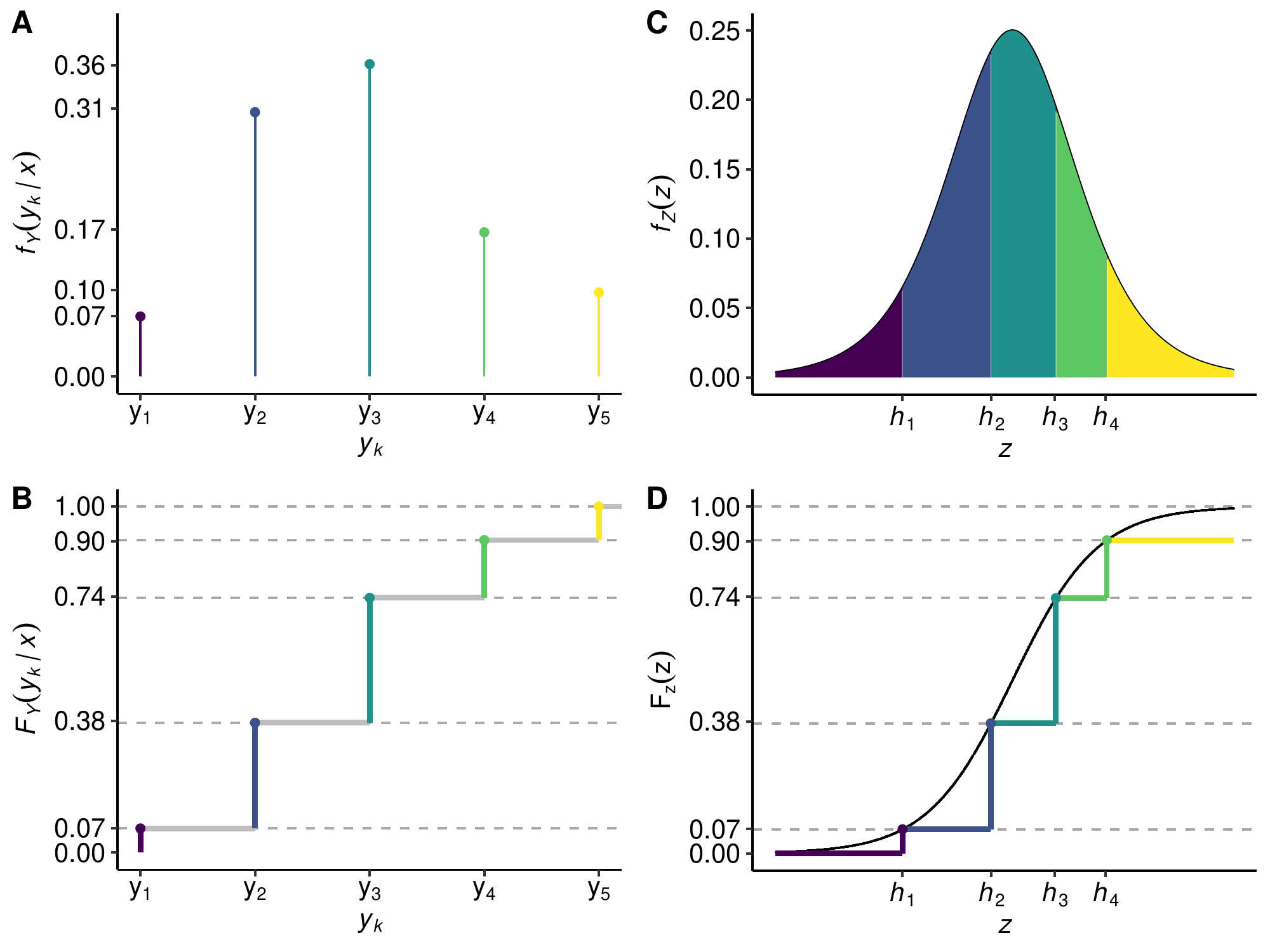}
\caption{
Density and distribution function for an ordered outcome with five classes.
(A) The probability density function (PDF) describes the probability of the
outcome belonging to class $k$. (B) The cumulative distribution function (CDF)
describes the probability of the outcome belonging to any class below and
including class $k$. The heights of the steps correspond to the probability of
belonging to class $k$, which are given by the PDF. Panels C and D depict the
PDF and CDF of the latent variable $Z$ and illustrate the equivalence between
modelling on the scale of $Y$ and $Z$. For brevity the cut points are denoted by
$\h_k := \h(\ry_k \given \rx)$. Note that $\h_5 = + \infty$ gives rise to the
yellow area corresponding to the conditional probability of $\rY$ belonging to the
fifth class, but does not show on the $x$-axis in panels C and D.
}\label{fig:ordinal}
\end{figure}

The distribution of an outcome variable is fully determined by its probability
density function (PDF). For an ordinal outcome the PDF describes the probabilities
of the different classes, which is equivalent to the PDF of an unordered
outcome. Yet, unlike an unordered outcome an ordered outcome possesses a well
defined cumulative distribution function (CDF) $\pY$, which naturally contains the order.
The CDF $\pY(\ry_k) = \prob(\rY \leq \ry_k)$ is a step function, which takes values
between zero and one and describes the probability of an outcome $Y$ of being smaller
or equal to a specific class. The steps are positioned at $\ry_k$, $k = 1,\dots,K$
and the heights of the steps correspond to the probability
$\prob(\rY = \ry_ k) = \pY(\ry_k) - \pY(\ry_{k-1}) =: \calL_i$ of observing
class $k$ (\emph{cf.}~Figure~\ref{fig:ordinal}~A and B). As indicated in the previous
subsection, the likelihood contribution for an observation $(\ry_{ki}, \rx_i)$ is given by the
predicted probability for the observed class, which can be rewritten as
\begin{align}
\calL_i = \prob(\rY=\ry_{ki} \given \rx_i)
  = \prob(\rY\leq\ry_{ki} \given \rx_i) - \prob(\rY\leq\ry_{(k-1)i} \given \rx_i),
\end{align}
for $k = 1,\dots,K-1$ and $\prob(\rY \leq \ry_0) := 0$, $\prob(\rY \leq \ry_K) = 1$.
Parametrizing the likelihood contributions using the CDF directly
enables to incorporate the order of the outcome when formulating
regression models for ordinal data (Section~\ref{sec:ontram}). 
It is worth noting that the loss is still the same negative log likelihood 
as in eq.~\eqref{eq:cce} merely using a different parametrization to take
the outcome's natural order into account. 

\begin{figure}[!ht]
\center
\includegraphics[width=0.7\textwidth]{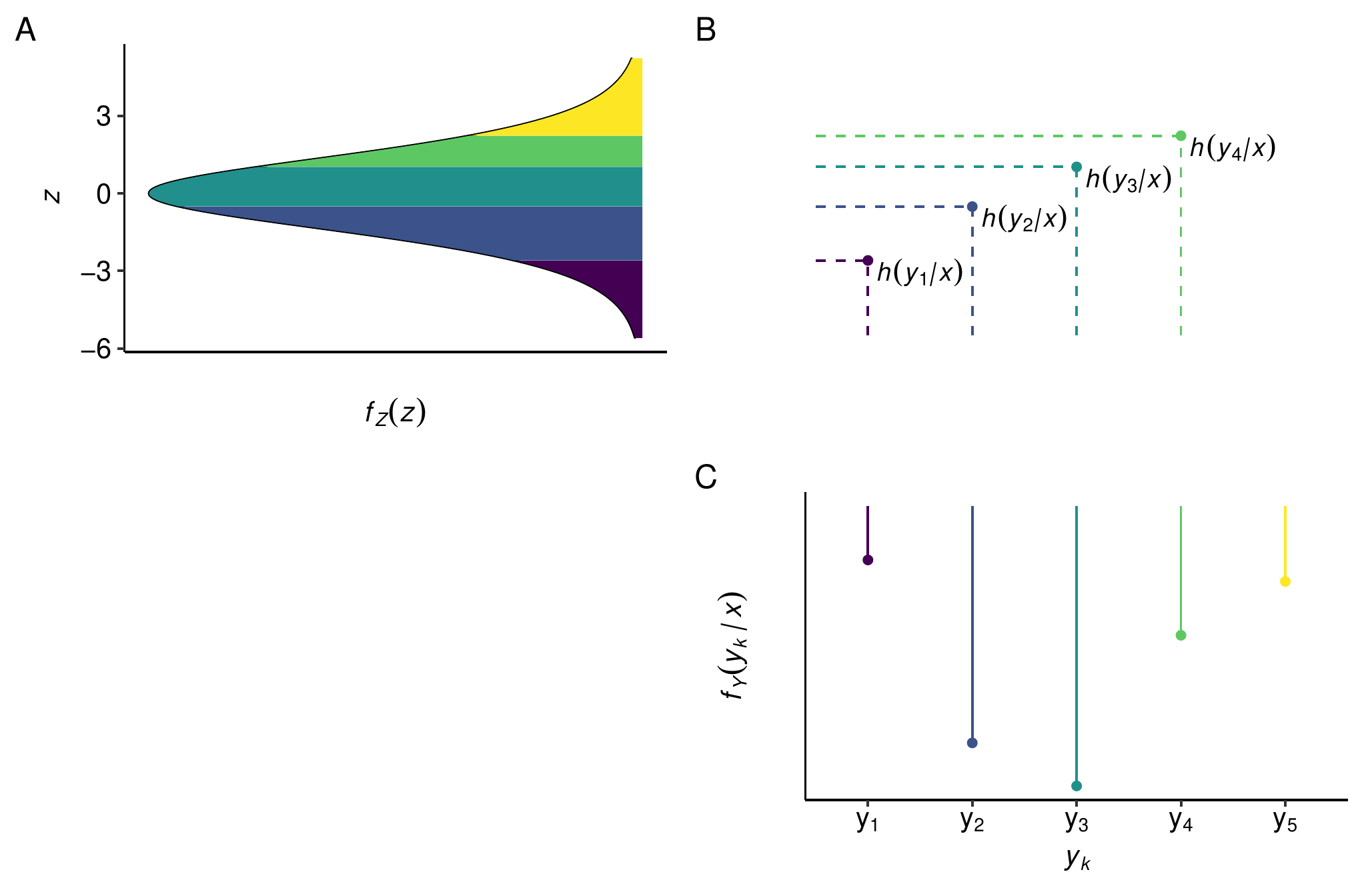}
\caption{
Transformation model likelihoods for a model with ordinal outcome.
Panel C shows the conditional density of $\rY$ given $\rx$, which gets mapped
onto the density of the latent variable $\rZ$ (A) via the transformation
function $\h$ (B). The likelihood contributions are in fact
probabilities and given by the area under the density of $\rZ$ between two consecutive
cut points in the transformation function. Note that $\h(\ry_5 \given \rx) = +\infty$
does not show on the plot for the transformation function, but is evident from the
yellow area under the density of $\rZ$.
}\label{fig:trafo}
\end{figure}

Many ordinal regression models assume the existence of an underlying continuous
latent variable (an unobserved quantity) $Z$. Sometimes field experts can give an
interpretation for the latent variable $Z$, e.g., the degree of illness of a patient.
The ordinal outcome variable $Y$ is understood as a categorized version of $Z$ resulting
from incomplete knowledge; we only know the classes in terms of the intervals in
which $Z$ lies. Fitting an ordinal regression model based on the latent variable
approach aims at finding cut points $\h(\ry_k \given \rx)$ at which $Z$ is separated into
the assumed classes (see Figure~\ref{fig:ordinal}~C). 
This is done by setting up a monotonically increasing step function 
$h$ that transforms the ordinal class values into cut points that retain the order 
(see Figure~\ref{fig:trafo}). 
Even if $Z$ can not be interpreted directly, using a latent variable approach has advantages,
because the chosen distribution of $Z$ determines the interpretability of the terms in the 
transformation function (see section \ref{sec:interpretation}). 

Moreover, the latent variable approach enables to understand ordinal regression as a 
special case of parametric transformation models, which were recently developed in statistics 
\citep{hothorn2014conditional} and are applicable to a wide range of outcomes
with natural extensions to classical
machine learning techniques such as random forests and boosting. 
Transformation models are able to model highly flexible outcome distributions while
simultaneously keeping specific model components interpretable. 
In transformation models the conditional outcome distribution of $(\rY\given\rx)$ is 
modeled by transforming the outcome variable $(\rY\given\rx)$ to a variable
$(\rZ\given\rx)$ with known (simple) CDF $\pZ$, like the Gaussian or logistic
distribution. Transformation models in general are thus defined by
\begin{align}
    \pY(\ry \given \rx)=\pZ(\h(\ry \given \rx)),
\end{align}
and all models in our proposed framework of {\sc ontram}s are of this form.

The first step to set up an ordinal transformation model is to choose a continuous 
distribution for $\rZ$, which determines the interpretational
scale of the effect of a predictor on the outcome (see~Section~\ref{sec:interpretation}).
The goal is then to fit a monotonically increasing transformation function $\h$, 
which maps the observed outcome classes $(\ry_k\given\rx)$ to the conditional cut points
\begin{align}
\h(\ry_k \given \rx), \; k = 1, \dots, K - 1,
\end{align}
of the latent variable $\rZ$, as illustrated in Figure~\ref{fig:trafo}. 
%
%
In the example in Figure~\ref{fig:trafo} the outcome can take five classes and
the $K-1$ cut points $\h(\ry_1\given\rx)$, $\h(\ry_2\given\rx)$, $\h(\ry_3\given\rx)$, 
and $\h(\ry_4\given\rx)$ have to be estimated. 
The first class of $\rY$ on the scale of $\rZ$ is given
by the interval $(-\infty, \h(\ry_1\given\rx)]$, the fifth class as $(\h(\ry_4\given\rx),+\infty)$,
so often the conventions $\h(\ry_0\given\rx) = - \infty$ and $\h(\ry_K\given\rx) = + \infty$
are used. The likelihood contribution of a given observation $(\ry_{ki},\rx_i)$
can now be derived from the CDF of $\rZ$ instead of $\rY$ and is given by
\begin{align}
\calL_i(\h;\ry_{ki}, \rx_i) &= \prob(\rY=\ry_{ki} \given \rx_i) \nonumber \\
  &= \pY(\ry_{ki} \given \rx_i) - \pY(\ry_{(k-1)i} \given \rx_i) \nonumber \\
  &= \pZ(\h(\ry_{ki} \given \rx_i)) - \pZ(\h(\ry_{(k-1)i} \given \rx_i)).
\end{align}
The single likelihood contributions are the heights of the steps in the CDF
or equivalently the area under the density of the latent variable $\rZ$
between two consecutive cut points (\emph{cf.} Figure~\ref{fig:trafo} B, C).
Note that two consecutive cut points enter the likelihood, such that the natural
order of the outcome is used to parametrize the likelihood, although the likelihood contribution
is given by the probability of the observed class alone.
Consequently, minimizing the negative log-likelihood
\begin{align}\label{eq:nll}
- \ell(\h;\ry_{1:n}, \rx_{1:n}) = - \sum_{i=1}^n \log \calL_i(\h;\ry_{ki}, \rx_i)
\end{align}
estimates the conditional outcome distribution of $(\rY\given\rx)$ by estimating
the unknown parameters of the transformation function, which in our application
are the cut points of $\rZ$. Note that in principle this formulation allows us to
directly incorporate uncertain observations, for instance, an observation may
lie somewhere in $[\ry_k,\ry_{k+2}]$, $k \leq K - 2$ if a rater is uncertain
about the quality of a wine or a patient rates their pain in between two classes.

\subsubsection{Interpretability in Proportional Odds Models} \label{sec:interpretation}

The interpretability of a transformation model depends on the choice of the
distribution $\pZ$ of the latent variable $\rZ$ and the transformation function $\h$.
A summary of common interpretational scales is given in Table~\ref{tab:interpr}.

\begin{table}[ht!]
\centering
\caption{
Interpretational scales of shift terms induced by $\pZ$ \citep{tutz2011regression}.
Most link functions have been studied in the context of proportional odds
model neural networks and a classification loss \citep{vargas2020cumulative}.
More details concerning the interpretational scales are given in 
Appendix~\ref{app:interpretation}.
}\label{tab:interpr}
\begin{tabular}{llll}
\toprule
\textbf{$F_Z$} & \textbf{$F_Z^{-1}$} & \textbf{Symbol} &
  \textbf{Interpretation of shift terms} \\ \midrule
Logistic & $\logit$ & $\pSL$ & log odds-ratio \\
Gompertz & cloglog & $\pMEV$ & log hazard-ratio \\
Gumbel & loglog & $\pGumbel$ & log hazard-ratio for $\rY_r = K + 1 - \rY$\\
Normal & probit & $\pN$ & not interpretable directly \\
\bottomrule
\end{tabular}
\end{table}

\begin{figure}[!ht]
\center
\includegraphics[width=0.9\textwidth]{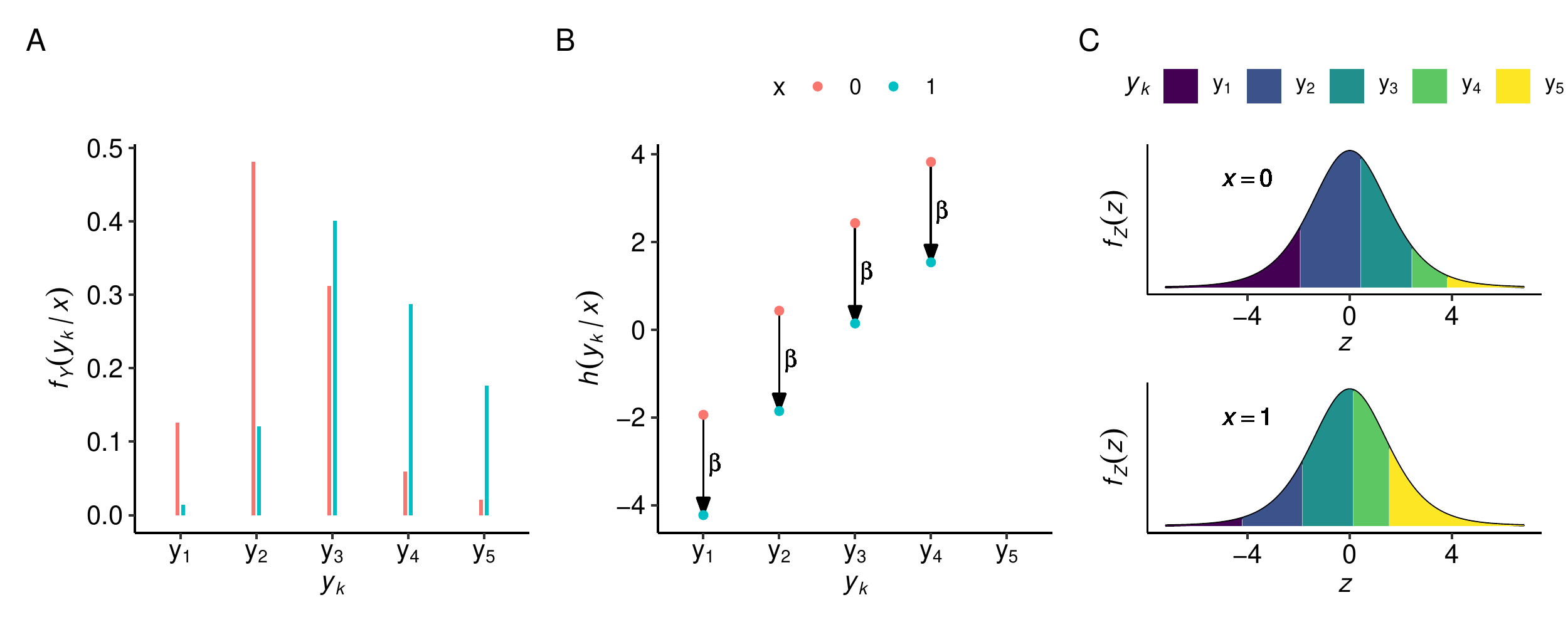}
\caption{
The conditional probability density, transformation function and latent representation
of the ordinal outcome $\rY$ with 5 classes depending on a single predictor $x$ which
is increased by $\Delta x = 1$ from 0 to 1. The density
of $(\rY \given x)$ for $x=0$ and $x=1$ is shown in A. The simple linear shift model
(see eq.~\eqref{eq:simp_shift}) imposes a downward shift of the transformation function by
$\beta$ when increasing the predictor from $x=0$ to $x=1$ (B). The shift in the
transformation function translates into a shift in the conditional cut points $\h(\ry_k\given\rx)$ under
the density of the latent variable $\rZ$ (C). Shifting the transformation
function downwards results in higher probabilities of $\rY$ belonging to a higher class.
}\label{fig:conditional}
\end{figure}

Here, we demonstrate interpretability through the example of a proportional odds model,
which is well known in statistics \citep{tutz2011regression}.
For the distribution of $\rZ$ we choose the standard logistic distribution
(denoted by $\pSL$), whose CDF is given by $\pZ(\rz) = \pSL(\rz) := (1 + \exp(-\rz))^{-1}$.
The transformation function $h$ is parametrized as
\begin{align} \label{eq:simp_shift}
\h(\ry_k \given \rx) = \eparm_k - \sum_{j=1}^J \eshiftparm_j x_j = \eparm_k - \linpred, 
    \; j = 1, \dots, J.
\end{align}
A transformation model with such a transformation function is called linear shift
model, since  a change $\Delta x_j$ in a single predictor $x_j$ causes a linear shift of size
$\beta_j \Delta x_j$ in the transformation function. In Figure~\ref{fig:conditional}
this is visualized for an outcome $\ry_k$ depending on a single predictor $x$, which is
increased by one unit (here from 0 to 1). This increase results in a simple shift
of size $\beta$ in the transformation function (Figure~\ref{fig:conditional}~B). 
However, the resulting conditional
distribution changes in a more complex way (Figure~\ref{fig:conditional}~A).
Note that the shape of the transformation function $h$ does not depend on $x$ and stays
unchanged while it is shifted downwards.
The popularity of this transformation model with $\pZ = \pSL$ is due to the insightful 
interpretation of the parameter $\eshiftparm_j$ as demonstrated in the following.

Based on the simple distribution $\pZ$ and the transformation function $\h$, 
the odds for the outcome to belong to a class higher than class $y_k$ can be written as
\begin{align}
\odds(\rY > \ry_k \given \rx)
&= \frac{\prob(\rY > \ry_k \given \rx)}{\prob(\rY \leq \ry_k \given \rx)}
= \frac{1-\pY(\ry_k \given \rx)}{\pY(\ry_k \given \rx)} \nonumber \\
&= \frac{1-\pZ(\h(\ry_k \given \rx))}{\pZ(\h(\ry_k \given \rx))}
= \frac{1-\pSL(\eparm_k - \linpred)}{\pSL(\eparm_k - \linpred)}.
\end{align}
When increasing the predictor $x_j$ by one unit and holding all other predictors
constant we change the vector $\rx$ to $\rx'$ and obtain, after some basic mathematical
transformation (\emph{cf.}~Appendix~\ref{app:interpretation}),
\begin{align}
\odds(\rY > \ry_k \given \rx')
&= \frac{1-\pSL(\eparm_k - \rx'^\top\shiftparm)}{ \pSL(\eparm_k - \rx'^\top\shiftparm)}
= \frac{1-\pSL(\eparm_k - \linpred - \eshiftparm_j)}{ \pSL(\eparm_k - \linpred - \eshiftparm_j)} \\
&= \odds(\rY > \ry_k \given \rx) \cdot \exp(\eshiftparm_j) \nonumber.
\end{align}
Note that the odds change by a constant factor $\exp(\eshiftparm_j)$ independent of $k$. 
Hence, the parameter $\eshiftparm_j$ in the linear shift term can be interpreted as the 
log-odds ratio of the outcome belonging to a higher outcome class than $\ry_k$ when 
increasing the predictor $x_j$ by one unit and holding the remaining predictors constant with 
\begin{align}
\log\OR_{\rx\to\rx'}
= \log\left(\frac{\odds(\rY > \ry_k \given \rx')}{\odds(\rY > \ry_k \given \rx)}\right)
= \eshiftparm_j.
\end{align}
This is depicted in Figure~\ref{fig:conditional} for a positive valued $\eshiftparm$, where the effect
of increasing the corresponding feature by one unit increases the odds for the outcome to belong to a 
higher class. More specific, the odds of the outcome for a higher class than $y_k$ is increased by a 
factor $\exp(\eshiftparm_j)$, which holds for each $y_k$. 
Because the effect of $\eshiftparm$ is the same for each class boundary these models are referred to
as proportional odds models \citep{tutz2011regression}. 

\subsection{Related Work} \label{sec:related}

Prediction models for ordinal outcomes have been studied in machine learning
as extensions of different popular methods like Gaussian Processes \citep{chu2005gaussian}, 
support vector machines \citep{chu2007support}, and neural networks \citep{cardoso2007learning}.
With the advent of deep learning, various approaches have been proposed to tackle
classification and regression tasks with ordinal outcomes, which we describe in more detail
in the following.
Note that we refer to models, which aim to predict a valid entire conditional outcome distribution as
ordinal \emph{regression} models, whereas models, which focus on the predicted class label
will be referred to as ordinal \emph{classification} models.
For instance, the MCC model described in Section~\ref{sec:background} is a 
regression model (\ie multinomial regression), whereas most of the state-of-the-art 
approaches described below are ordinal classifiers.
In the following we discuss literature on ordinal classification and literature related 
to different aspects of our work, \ie regression models, transformation models,
and interpretability.

\paragraph{Ordinal classification}
Deep learning approaches to ordinal regression and classification problems 
range from using an ordinal metric for the evaluation of multi-class
classification models to the construction of novel ordinal loss functions and
dummy encodings.
The earliest approaches made use of the equivalence of an ordinal prediction problem
with outcome $\rY \in \{\ry_1<\dots<\ry_K\}$, to the $K-1$ binary classification problems
given by $\1(\rY\leq\ry_k)$, $k = 1,\dots,K$ \citep{frank2001simple},
which is still being used in applications such as age estimation
\citep{niu2016ordinal}.

\citet{cheng2008neural} devised a cumulative dummy encoding for the
ordinal response where for $\rY=\ry_k$ we have $\ry_i = 1$ if $i\leq k$ and 0
otherwise.
\citeauthor{cheng2008neural} then suggest a sigmoid activation for the last layer
of dimension $K$, together with two loss functions (relative entropy and a squared error loss).
Similar approaches remain highly popular in application.
For instance, \citet{cao2019rank} extend the approach to rank-consistent
ordinal predictions. 
The problem of rank inconsistency, however, is confined to the $K$-rank and similar
approaches and does not appear in ordinal regression models, such as the ones we propose.

\citet{xie2019deep} used a similar dummy encoding to train $K-1$ binary classifiers,
which share a common CNN trunk for image feature extraction but possess their own
fully connected part. This allows flexible feature extraction while reducing model
complexity substantially due to weight sharing.
Weight sharing is a natural advantage of models which are trained with an ordinal
loss function instead of multiple binary losses, which we describe next.
A comparison of {\sc ontram} against the method described in \citet{xie2019deep}
can be found in Appendix~\ref{app:resqwk}.

Recently, the focus shifted towards novel ordinal loss functions involving
Cohen's kappa, which was first proposed by \citep{de2018weighted} and
subsequently used in ``proportional odds model (POM) neural networks''
\citep{vargas2019deep}.
POM neural networks and their extensions to other cumulative link functions
in \citep{vargas2020cumulative} are closely related to {\sc ontram}s, proposed 
in this paper, because they constitute a special case in which the class-specific
intercepts do not depend on input data (see Section~\ref{sec:ontram}).
The crucial difference between POM NNs (as proposed in \citep{vargas2019deep})
and {\sc ontram}s is the quadratic weighted Cohen's kappa (QWK) loss function 
in POM NNs, compared to a log-likelihood loss in {\sc ontram}s.
Although POM NNs predict a full conditional outcome distribution, their
focus lies on optimizing a classification metric (QWK). 
The idea is to penalize misclassifications that are further away from the 
observed class stronger than misclassifications that are closer to the observed class.
In contrast, in regression approaches, the goal is to predict a valid probability 
distribution across all classes.
We give more detail on and compare our proposed method against the QWK loss
in Appendices~\ref{app:qwk}~and~\ref{app:resqwk}, respectively.
We use QWK-based models as an example to address the general problem arising when 
comparing classification and regression models, which address different questions
and hence optimize distinct target functions.

\paragraph{Ordinal regression}
Lastly, \citet{liu2019probabilistic} took a probabilistic approach using Gaussian processes
with an ordinal likelihood similar to the cumulative probit model (cumulative ordinal model 
with $\pZ=\pN$) and a model formulation similar to POM neural networks.
We address further related work concerning technical details in 
Section~\ref{sec:ontram}, such as the explicit formulation of constraints in the
loss function.

\paragraph{Transformation models}
Deep conditional transformation models have very recently been applied to regression
problems with a continuous outcome \citep{sick2020deep, baumann2020deep}.
Both approaches, as well as our contribution, can be seen as special cases of
semi-structured deep distributional regression, which was proposed by \citet{rugamer2020unifying}.
\citeauthor{sick2020deep} parametrized the transformation function as a composition of
linear and sigmoid transformations and a flexible basis expansion that ensures
monotonicity of the resulting transformation function.
The authors applied deep transformation models to a multitude of benchmark data sets
with a continuous outcome and demonstrated a performance that was comparable to
or better than other state-of-the-art models. However, in one of the benchmark
data set the authors treated a truly ordinal outcome as continuous, as done by
all the other benchmark models. This is indicative for the lack of deep learning
models for ordered categorical regression.

\paragraph{Interpretability}
In general, deep learning models suffer from a lack of interpretability
of the predictions they make \citep{goodfellow2016deep}. 
In DL models related to image data, interpretability is mostly referred to as
highlighting parts of the image that explain the respective prediction. 
Often, surrogate models are build on top of the black-box model's
predictions, which are easier to interpret. 
One such model is LIME \citep{ribeiro2016should}.
For problems with an ordinal outcome, 
\citet{amorim2018interpreting} comment on the limited interpretability of the
ensemble of neural networks in the $K$-rank approach described above and propose to use
a mimic learning technique, which combines the ensemble with a more directly
interpretable model.
In the present work we take a different approach to interpretability rooted in statistical
regression models. 
The interpretability of the effect of individual input features is given
by the fitted model parameters in an additive transformation function,
which is a common modelling choice for achieving interpretability \citep{rudin2018stop}.
\citet{agarwal2020neural} take the same approach in the framework of generalized additive models (GAMs).
We give more detail in Section~\ref{sec:interpretation} and Appendix~\ref{app:interpretation}.

\section{Ordinal Neural Network Transformation Models} \label{sec:ontram}

\begin{figure}[!ht]
\center
\includegraphics[width=0.85\textwidth]{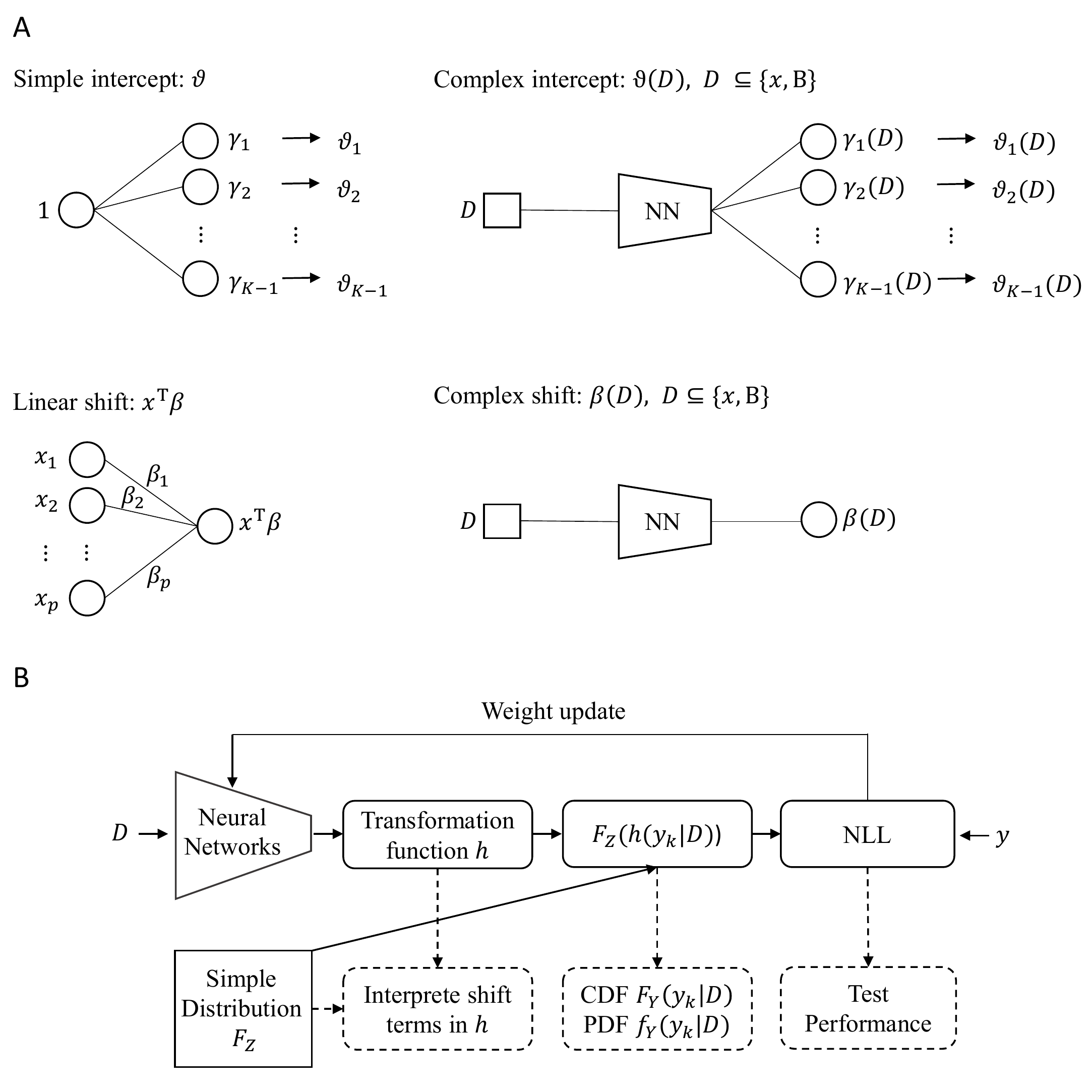}
\caption{
Architecture building blocks of {\sc ontram}s. A: Simple intercept and 
linear shift terms are modeled using a single-layer neural network.
Complex intercept and complex shift terms are allowed to depend on the 
input data in a more complex manner and may be a fully connected or 
convolutional NN depending on the type of the input data.
The input data $D$ can be a single tabular predictor $x_j$ or a set of 
tabular predictors $\rx$ or images $\B$.
B: The output of the NNs control the additive components of the 
transformation function $h$.
Together with the choice of $\pZ$, $h$ determines the full model,
from which the likelihood can be evaluated.
During training time (solid lines) the weights of all model components 
are trained jointly by minimizing the NLL. 
After training (dashed lines) the shift terms in the transformation function 
can be interpreted, the conditional outcome distribution can be predicted and 
the NLL can be evaluated for a given test set.
}\label{fig:ONTRAMarchitecture}
\end{figure}

Here, we present ordinal neural network transformation models, which unite cumulative
ordinal regression models with deep neural networks and seamlessly integrate complex
data like images ($\B$) and/or tabular data ($\rx$). At the heart of an {\sc ontram} lies a
parametric transformation function $\h(\ry_k\given \rx, \B)$, which transforms the ordinal
outcome $y_k$ to cut points of a continuous latent variable and
controls the interpretability and flexibility of the model (see Figure~\ref{fig:trafo}).
The ordering of the outcome is incorporated in the {\sc ontram} loss function by defining
it via the cumulative distribution function
\begin{align}\label{ontram_nll}
\text{NLL} := 
    - \frac{1}{n} \sum_{i=1}^n \log
    \left(
      \pZ(\h(\ry_{ki} \given \rx_i, \B_i)) - \pZ(\h(\ry_{(k-1)i} \given \rx_i, \B_i))
    \right).
\end{align}
In the following we describe the terms of the parametric transformation function
and their interpretability. The parameters of these terms are controlled by NNs,
which are jointly fitted in an end-to-end fashion by minimizing the NLL 
(Figure~\ref{fig:ONTRAMarchitecture}).

\paragraph{Modularity}
The transformation function, which determines the complexity and interpretability of an
{\sc ontram}, consists of an intercept term, optionally followed 
by additive shift terms, which depend in a more or less complex manner on different
input data and are controlled by NNs (see Figure~\ref{fig:ONTRAMarchitecture}).

The intercept term controls the shape of the transformation function:
\begin{enumerate}
  \item Simple intercepts (SI) $\eparm_k$, $k=1,\dots,K-1$ are unconditional, i.e. the
    shape of the transformation function is independent of the input data. SIs can be
    modeled as a single layer neural network with $K-1$ output units and linear
    activation function. The input is given by 1. The outputs are given
    by $\gamma_1,\dots,\gamma_{K-1}$ controlling the intercepts 
    (see Figure~\ref{fig:ONTRAMarchitecture}).
  \item Complex intercepts (CI), on the other hand, depend on the input data,
    which may be tabular data, image data or
    a combination of both, yielding $\eparm_k(\rx)$, $\eparm_k(\B)$, or $\eparm_k(\rx,\B)$, respectively.
    CIs enable more complex transformation functions, whose shape may
    vary with the input.
    Depending on the type of input data, CIs are modeled using a multi-layer fully 
    connected neural network, a convolutional neural network or a combination of both.
    Analogous to SI terms, the number
    of output units in the last layer is equal to $K-1$ with linear activation
    function, yielding $\gamma_1(\rx, \B), \dots, \gamma_{K-1}(\rx, \B)$ depending on the input
    (see upper right panel in Figure~\ref{fig:ONTRAMarchitecture}).
\end{enumerate}
To ensure that the transformation function is non-decreasing, the outputs
$\gamma_1, \dots, \gamma_{K-1}$ of simple and complex intercept models are transformed before
entering the likelihood via
\begin{align} \label{eq:constraints}
\begin{split}
  \eparm_k &= \eparm_1 + \sum_{i = 2}^k \exp(\gamma_i), \quad k = 2, \dots, K - 1, \\
  \eparm_0 &= - \infty, \; \eparm_1 = \gamma_1, \; \eparm_K = + \infty.
\end{split}
\end{align}
The addition of $\eparm_0 = - \infty$ and $\eparm_K = + \infty$ is
important for computing the loss as described in Section~\ref{sec:background}.
Enforcing a monotone increasing transformation function via eq.~\eqref{eq:constraints}, such that
$\eparm_0 < \eparm_1 \leq \dots < \eparm_K$, has been done similarly 
in the literature.
In what \citet{cheng2008neural} call threshold models, $\gamma_i$
is squared instead of taking the exponential to ensure the intercept function is
non-decreasing \citep{liu2019probabilistic,vargas2020cumulative}.
A different but related approach is to softly penalize the loss for pair-wise rank 
inconsistencies using a hinge loss \citep{liu2017deep,liu2018constrained}.
Note that the special case $\eparm_k(\rx,\B)$ already includes both tabular
and image data. That is, the transformation function and therefore the outcome
distribution is allowed to change with each input $\rx$ and $\B$, which represents
the most flexible model possible.
In fact, this most flexible {\sc ontram} is equivalent to a MCC model with softmax as 
last-layer activation function and a categorical cross-entropy loss, albeit parametrized 
differently to take the order of the outcome into account.

Shift terms impose data dependent vertical shifts on the transformation function:
\begin{enumerate}
  \item Linear shift (LS) terms $\linpred$ are used for tabular features
    and are directly interpretable (see Section \ref{sec:interpretation}).
    The components of the parameter $\shiftparm$ can be modeled as the weights
    of a single layer neural network with input $\rx$, one output unit with
    linear activation function and without a bias term
    (see lower left panel in Figure~\ref{fig:ONTRAMarchitecture}).
  \item Complex shift (CS) terms depend on tabular predictors or image data.
  Complex shift terms are modeled using flexible dense and/or convolutional
  NNs with input $\rx$ and/or $\B$, and a single output unit with linear activation
  (see lower right panel in Figure~\ref{fig:ONTRAMarchitecture}). Similar to
  linear shift terms, the output of $\beta$ and $\eta$ can be interpreted as the log odds
  of belonging to a higher class, compared to all lower classes, if $\pZ=\pSL$.
  Again, this effect is common to all class boundaries. In contrast
  to a linear shift term, we can model a complex shift for each tabular 
  predictor $\beta(x_j)$ akin to a generalized additive model. Alternatively, we can model
  a single complex shift $\beta(\rx)$ for all predictors, which allows for higher order 
  interactions between the predictors.
  This way, the interpretation of an effect of a single predictor is lost in favour
  of higher model complexity.
\end{enumerate}

\paragraph{Interpretability and flexibility}
In the following, we will present a non-exhaustive
collection of {\sc ontram}s integrating both tabular and image data. We start to
introduce the least complex model with the highest degree of interpretability
and end with the most complex model with the lowest degree of interpretability.

The simplest {\sc ontram} conditioning on tabular data $\rx$ and image data $\B$ is given by
\begin{align}
\h(\ry_k \given \rx, \B) = \eparm_k - \linpred - \eta(\B),
\end{align}
where $\eparm_k$ is a simple intercept corresponding to class $k$, $\shiftparm$
is the weight vector of a single layer NN as described above and $\eta(\B)$ the
output of a CNN (Figure~\ref{fig:ONTRAMarchitecture}~A).
In this case, $\shiftparm$ and $\eta$ can be interpreted as cumulative log odds-ratios
when choosing $\pZ = \pSL$ (see Section~\ref{sec:interpretation}).
The above model can be made more flexible, yet less interpretable, by substituting
the linear predictor for a more complex neural network $\beta$, such that
\begin{align}
\h(\ry_k \given \rx, \B) = \eparm_k - \beta(\rx) - \eta(\B),
\end{align}
where $\beta(\rx)$ is now a log odds ratio function that allows for higher order
interactions between all predictors in $\rx$. For instance, one may be interested
in the odds ratio $\OR_{\B \to \B'}$ of belonging to a higher category when changing
an image $\B$ to $\B'$ and holding all other variables constant.
As a special case, complex shifts include an additive model formulation in the spirit
of generalized additive models (GAMs) by explicitly parametrizing the effect of each
predictor $x_j$ with a single neural network $\eshiftparm_j$ 
\citep{agarwal2020neural}
\begin{align} \label{eq:gam_shift}
\h(\ry_k \given \rx, \B) = \eparm_k - \sum_{j=1}^J \eshiftparm_j(x_j) - \eta(\B), \; j = 1, \dots, J.
\end{align}
For $\pZ = \pSL$ the complex shift term $\eshiftparm_j(x)$ can be interpreted as a log-odds ratio 
for the outcome to belong to a higher class than $y_k$ compared to the scenario where 
$\eshiftparm_j(x)=0$, all other predictors kept constant.

Another layer of complexity can be added by allowing the intercept function $\eparm_k$
for $\rY = \ry_k$, to depend on the image
\begin{align}
\h(\ry_k \given \rx, \B) = \eparm_k(\B) - \beta(\rx).
\end{align}
In this transformation function we call $\eparm_k(\B)$ complex intercept, because the
intercept function is allowed to change with the image (Figure~\ref{fig:ONTRAMarchitecture}~A).
One does not necessarily have to stop here.
Including both the image and the tabular data in a complex intercept
\begin{align}
\h(\ry_k \given \rx, \B) = \eparm_k(\rx, \B)
\end{align}
represents the most flexible model whose likelihood is equivalent to the
one used in MCC models, solely with a different parametrization.
Consequently, solely the most flexible {\sc ontram}s achieve on-par performance
compared with deep classifiers trained using the cross-entropy loss, while the
less flexible {\sc ontram}s are attractive because of their easier interpretability.
In fact, we illustrate empirically that a minor trade-off in predictive performance leads to a
considerable ease in interpretation.

\paragraph{Computational details}
The parameters of an {\sc ontram} are jointly trained via stochastic gradient descent.
The parameters enter the loss function via the outputs of the simple/complex intercept and
shift terms modeled as neural networks (see Figure~\ref{fig:ONTRAMarchitecture}~A). 
The gradient of the loss with respect to all trainable parameters is computed via 
automatic differentiation in the \texttt{TensorFlow} framework. 
Note that any pre-implemented optimizer can be used and that there are no constraints on the
architecture of the individual components besides their last-layer dimension and activation
function.

\section{Experiments} \label{sec:experiments}

We perform several experiments on data with an ordinal outcome to evaluate and
benchmark {\sc ontram}s in terms of prediction performance and interpretability.
For the experiments we use two publicly available data sets as presented in
the following section. In addition, we simulate tabular predictors to assess
estimation performance for the effect estimates in {\sc ontram}s.

\subsection{Data} \label{sec:data}

\begin{figure}[!ht]
\centering
\includegraphics[width=\textwidth]{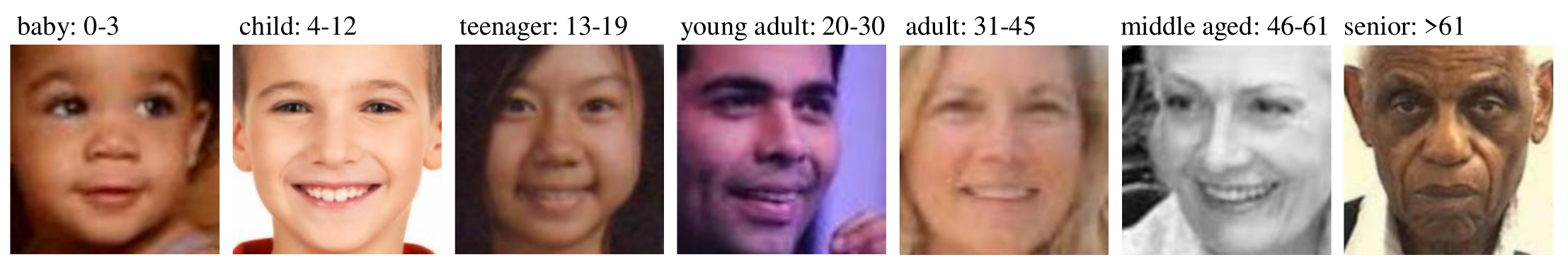}
\vspace{-0.8cm}
\caption{
Example images for UTKFace. Example images of
the seven ordinal age-classes (baby, child, teenager, young adult, adult, middle
aged and senior) of the cropped and aligned UTKFace data set are presented.
} \label{fig:images}
\end{figure}

\paragraph{UTKFace}
UTKFace contains more than 23000 images of faces belonging
to all age groups \citep[dataset][]{utkface}. The ordinal outcome is determined by age using the
classes baby (0--3, $n_0 = 1894$), child (4--12, $n_1 = 1519$), teenager (13--19, $n_2 = 1180$),
young adult (20--30 $n_3 = 8068$), adult (31--45, $n_4 = 5433$),
middle aged (46--61, $n_5 = 3216$) and senior ($>$61, $n_6 = 2395$) \citep[dataset][]{das2018}. 
The images are labeled with the people’s age (0 to 116) from which the age-class is determined.
In addition, the data set provides the tabular feature sex (female, male).
As our main goal is not on performance improvement but on the evaluation of our
proposed method, we use the already aligned and cropped versions of the 
images. For some example images see Figure~\ref{fig:images}.

\begin{figure}[!ht]
\centering
\resizebox{0.6\textwidth}{!}{%
\begin{tikzpicture}[auto, Arr/.style={-{Latex[length=1.5mm]}}]
  \node (X2) {$X_{\{2, 3\}}$};
  \node (Xn) [below =1.5cm of X2] {$X_{\{1, 4, 7, 10\}}$};
  \node [below =1.5cm of Xn] (X5) {$X_{\{5, 6\}}$};
  \node [below =1.5cm of X5] (X6) {$X_{\{8, 9\}}$};
  \node [right =8cm of Xn] (Y) {$Y$};
  \node [right =3cm of Y] (B) {$\B$};
  \draw[Arr] (X2) to [bend left=20, "$\pm \log 2$"] (Y);
  \draw[Arr] (Xn) to ["$0$"] (Y);
  \draw[Arr] (X5) to [bend right=20, "$\pm \log 1.5$"] (Y);
  \draw[Arr] (X6) to [bend right=20, "$\pm \log 3$"] (Y);
  \draw[Arr, dotted] (B) to (Y);
\end{tikzpicture}
}
\caption{Simulation of predictors for UTKFace data.
$X_j \stackrel{\iid}{\sim} \ND(0, \sigma^2)$, $j = 1, \dots, 10$,
with $\sigma^2 = 1.55^2$.
The predictors $X_j$ are simulated such that their effects adhere to the proportional 
odds assumption. 
That is, the effect of $\shiftparm$ is common to all class boundaries. Note that the arrows
indicate effects on the log-odds scale of the outcome $\rY$, i.e.,
$\pY(\ry_k \given \rx) = F_L(\eparm_k - \linpred)$. The dotted arrow from
$\B$ to $\rY$ indicates that the image is not entering the simulation directly but
is assumed to have an effect on the outcome.} \label{fig:sim}
\end{figure}
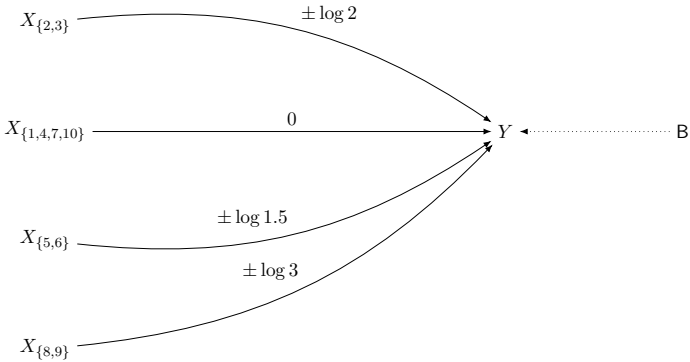

We simulate tabular predictors $\rx$ with predefined effects on the ordinal
outcome of the UTKFace data set, where we assume a proportional odds model
$\pY(\ry_k \given \rx) = \pZ(\eparm_k - \linpred)$ (see Section \ref{sec:interpretation}).
The simulation scheme is closely related to choice-based sampling \citep{manski1977estimation}.
Ten predictors are simulated, four of which are noise predictors
that have no effect on the outcome. The six informative predictors are simulated
to have an effect of $\pm \log 1.5$, $\pm \log 2$ and $\pm \log 3$ on the log-odds scale,
to reflect small to large effect sizes commonly seen in medical and epidemiological
applications.
All predictors are mutually independent of each other and the image data.
The predictors are simulated from the conditional distribution 
$X_j \given \boldsymbol{y} \sim \calN(\boldsymbol{y}^\top\boldsymbol\xi_j, \sigma^2)$, where
$\boldsymbol{y}$ denotes the one-hot encoded outcome and $(\boldsymbol\xi_j)_k = k \cdot \beta_j$
for $k = 1, \dots, K$, to emulate the proportional odds property.
All simulated predictors have a common variance of $\sigma^2 = 1.55^2$, which was tuned
such that the estimated $\hat\beta_j$ on average recovers the true $\beta_j$ when 
estimating the conditional distribution $\rY \given \rx$ in a proportional odds model.
A summary of the simulation procedure is given in Figure~\ref{fig:sim}.

\paragraph{Wine quality}
The Wine quality data set consists of 4898 observations \citep[dataset][]{cortez2009modeling}.
The ordinal outcome describes the wine quality measured on a scale with 10 levels
of which only 6 consecutive classes (3 to 8, $n_3 = 10$,  $n_4 = 53$,  $n_5 = 681$,
$n_6 = 638$,  $n_7 = 199$,  $n_8 = 18$) are observed.
The data set contains 11 predictors, such as acidity, citric acid and sugar content.
As in \citet{gal2015dropout}, we consider a subset of the data (red wine, $n = 1599$).

\subsection{Models} \label{sec:models}
The models we use for evaluating and benchmarking the proposed {\sc ontram}s are
summarized in Table~\ref{tab:models}.
\begin{table}[!ht]
\centering
\caption{%
Summary of the models used for evaluating the {\sc ontram} methods.
In the upper part we list models used for the Wine data, which contain only
tabular predictors ($\rx$). In the lower part, we show models for the UTKFace data, which consist
of image data and tabular predictors $(\rx,\B)$. Above the thin lines we list the
baseline models; below the {\sc ontram}s. For each model, which can be framed as a
transformation model, the transformation function is given. Parameters in the shift
terms of a transformation function can be interpreted as log odds-ratios if $F_Z$ is
chosen to be the standard-logistic distribution. 
Then, any model involving a simple intercept is an instance of a proportional odds 
model.} \label{tab:models}
\resizebox{\textwidth}{!}{%
\begin{tabular}{@{}llll@{}}
\toprule
\textbf{Data set}
& \textbf{Model name}             & \textbf{Abbreviation} & \textbf{Trafo} $\h(\ry_k \given \rx, \B)$ \\ \midrule
\multirow{8}{*}{UTKFace}
& Multi-class classification       & MCC                   &                \\
& Multi-class classification + tabular& MCC-$\rx$            &                \\
\cline{2-4}
& Complex intercept               & CI$_\B$               & $\eparm_k(\B)$ \\
& Complex intercept + tabular     & CI$_\B$-LS$_{\rx}$        & $\eparm_k(\B) - \linpred$ \\
& Simple intercept + complex shift& SI-CS$_\B$            & $\eparm_k - \eta(\B)$ \\
& Simple intercept + complex shift + tabular& SI-CS$_\B$-LS$_{\rx}$ & $\eparm_k - \eta(\B) - \linpred$ \\ 
& Simple intercept + tabular      & SI-LS$_{\rx}$           & $\eparm_k - \linpred$     \\
\midrule
\multirow{5}{*}{Wine}
& Multi-class classification       & MCC                   &                \\
& Generalized additive proportional odds model   & GAM  &  $\eparm_k - \sum_{j=1}^p \eshiftparm_j(x_j)$ \\ 
& Proportional odds logistic regression & polr                  &  $\eparm_k - \linpred$ \\
\cline{2-4}
& Complex intercept               & CI$_{\rx}$                &  $\eparm_k(\rx)$ \\
& Simple intercept + GAM complex shift   & SI-CS$^\star_{\rx}$  &  $\eparm_k - \sum_{j=1}^p \eshiftparm_j(x_j)$ \\ 
& Simple intercept + linear shift & SI-LS$_{\rx}$             &  $\eparm_k - \linpred$ \\ \bottomrule
\end{tabular}}
\end{table}
These models feature a different flexibility
and interpretability and are trained with the different loss functions described
in Sections~\ref{sec:background}~and~\ref{sec:ontram}.

\subsection{Training and Validation Setup} \label{sec:expsetup}

The models are implemented and evaluated on the UTKFace and Wine quality data set as 
described in the following. 
The NN architectures used for the different applications are summarized in Appendix~\ref{app:NN}.

\paragraph{UTKFace}
The UTKFace data set is split into a training (80\%) and a test set (20\%);
20\% of the training set is used as validation data. The images are resized to 
$128\times128\times3$ pixels.
The resized images are normalized to have pixel values between 0 and 1.
No further preprocessing is performed.
The models are trained for up to 100 epochs with the Adam optimizer with
a batch size of 32 and a learning rate of 0.001.
Overfitting is addressed with early stopping. That is, we retrospectively search
for the epoch in which the validation loss is minimal and consider the
respective parameters for our trained models.

We analyse the data set using deep ensembling \citep{lakshminarayanan2017simple}, 
a state of the art approach in probabilistic deep learning methods leading to more 
reliable probabilistic predictions \citep{wilson2020bayesian}.
Specifically, models are trained five times with a different weight 
initialization in each iteration. 
The resulting predicted conditional outcome distribution is averaged over the five runs and
this averaged conditional outcome distribution is then used for model evaluation.
This procedure is supposed to prevent double descent and
improve test performance \citep{wilson2020bayesian}.

\paragraph{Wine quality}
For analyses of the wine quality data we employ the same cross-validation
scheme as \citet{gal2015dropout} and split the data into 20 folds of 90\%
training and 10\% test data.
The predictors are normalized to the unit interval. 
Otherwise, no further preprocessing is performed.
The architectures of the used NNs are described in Appendix \ref{app:NN}.
All NNs are jointly trained via stochastic gradient descent using a learning rate of 0.001.
The linear shift model is trained for 8000 epochs with a batch size of 90.
The {\sc ontram} GAM is trained for 1200 epochs with a batch size of 180.
All other models  are trained for 100 epochs with 
a batch size of 6.

\subsection{Software} \label{sec:software}

We implement MCC models and {\sc ontram}s in the two programming languages \proglang{R} 3.6-3
and \proglang{Python} 3.7.
The models are written in \texttt{Keras} based on a \texttt{TensorFlow} backend using
\texttt{TensorFlow} version $>$2.0 \citep{chollet2015keras,tensorflow2015} and trained
on a GPU.
Both polr and generalized additive proportional odds models are fitted in \proglang{R}
using \cmd{tram::polr} \citep{pkg:tram} and \cmd{mgcv::gam} \citep{wood2017gam}, respectively.
Further analysis and visualization is performed in \proglang{R}.
For reproducibility, all code is made available on 
GitHub.\footnote{\url{https://www.github.com/LucasKookUZH/ontram-paper}}

\subsection{Model Evaluation} \label{sec:eval}

\emph{Evaluation metrics:}
The main focus of {\sc ontram}s is to be able to interpret their individual components
and the most flexible {\sc ontram} is equivalent to the MCC model.
In turn, prediction performance of {\sc ontram}s can only ever be as good as in MCC.
Therefore, we assess prediction performance mainly to illustrate trading off model 
flexibility against ease of interpretation.
We evaluate the prediction performance of {\sc ontram}s and MCCs with proper scoring rules,
namely the negative log-likelihood (NLL) and the ranked probability score (RPS).
Roughly speaking, proper scoring rules encourage honest probabilistic predictions
because they take their optimal value when the predicted conditional outcome
distribution corresponds to the data generating distribution
(for details see Appendix~\ref{app:scoring}). 
In Appendix~\ref{app:confmats} we compute additional evaluation metrics which are 
commonly used for ordinal classification models, \ie accuracy and QWK which is 
discussed in Appendix~\ref{app:qwk}.

\paragraph{Estimation and interpretability}
To evaluate whether {\sc ontram}s yield reliably interpretable effect estimates of shift 
components we make use of the simulated tabular predictors and compare the known
true effects of the individual predictors to the estimates.
For other predictors we discuss the plausibility of the estimated effects or, 
if applicable, compare them to results of other benchmark experiments.

\section{Results} \label{sec:discussion}

Results for the MCC models and {\sc ontram}s for the UTKFace and wine data
are given in the following Sections~\ref{sec:resdrface} and \ref{sec:reswine},
respectively.

\subsection{UTKFace} \label{sec:resdrface}

We first evaluate {\sc ontram}s on the UTKFace data set, which contains images
and tabular predictors that allow to illustrate the interpretation of the shift terms.
As in other applications, age is discretized and treated as an ordinal outcome 
(see \eg \citep[]{liu2017ordinal}).

We first train a SI-CS$_{\B}$-LS$_{\text{sex}}$ {\sc ontram} with transformation function 
$\h(\ry_{k} \given \rx)=\eparm_k - \eta(\B) - \beta_{\text{sex}}\cdot \I(\text{sex}=\text{female})$
that includes the tabular predictor sex in addition to the images.
We assume that the prediction of the age class depends on the appearance of a person and 
therefore on the image but not on a person's sex.
On the other hand, a person's sex can often be deduced from an image, 
which renders the tabular feature and image data collinear and makes 
estimation and interpretability of the individual effects more difficult.
However, collinear data is representative for most practical applications. 
We thus expect the estimated coefficient $\beta_{\text{sex}}$ 
to be small in comparison to the effect of the image $\eta(\B)$, 
which we expect to be a better predictor of a person's age.

\begin{figure}[!ht]
\centering
\includegraphics[width=0.55\textwidth]{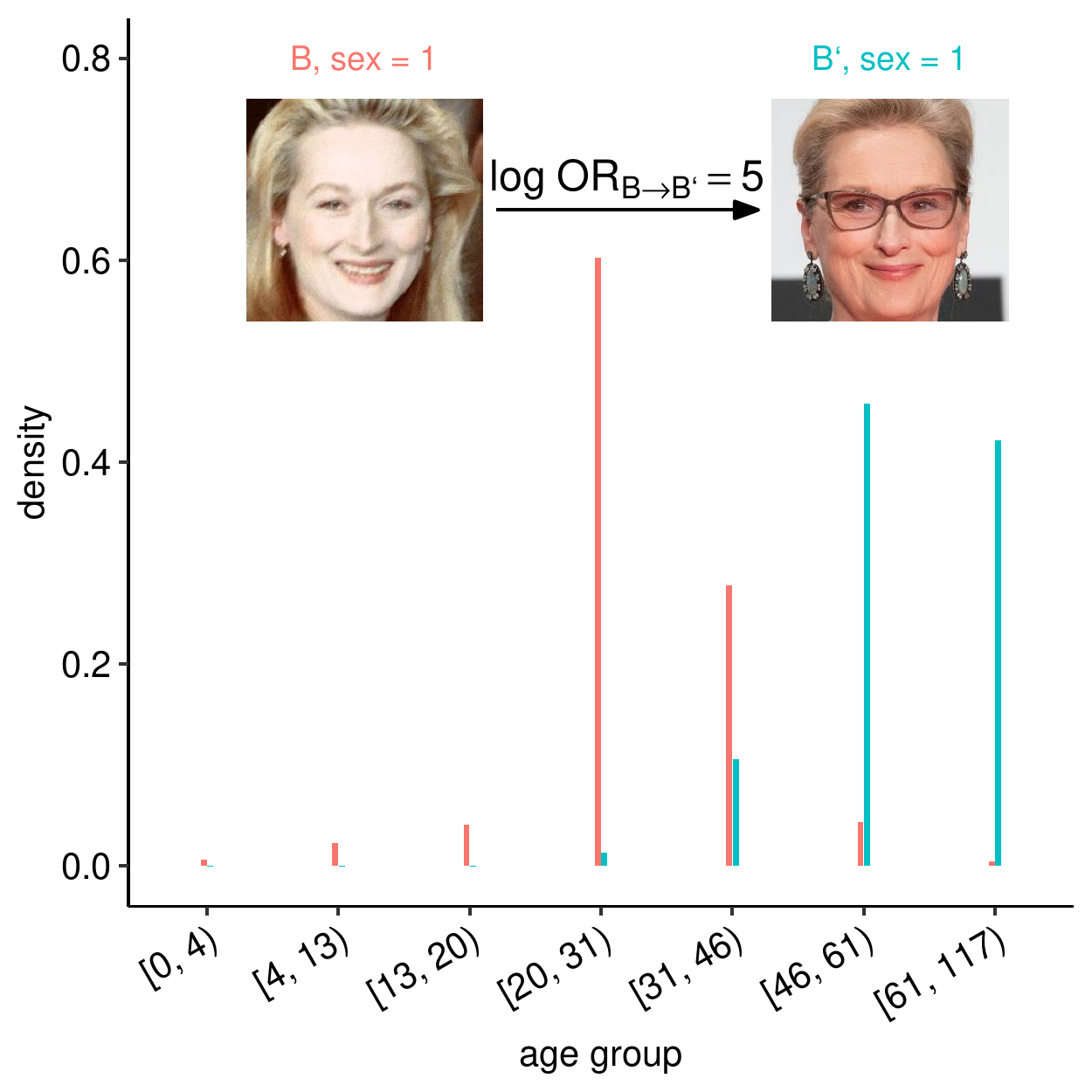}
\caption{
Predicted densities in a SI-CS$_{\B}$-LS$_{\text{sex}}$ {\sc ontram} once 
using the image of a 41 year-old and a 67 year-old Meryl Streep, while
the gender is female in both cases.
What sets {\sc ontram}s apart from other ordinal DL classifiers, is the
directly interpretable effect of changing an image in terms of a log odds-ratio.
Namely, the odds of belonging to a higher age change by a factor of $\exp(5)$ 
when changing image $\B$ to $\B'$, keeping sex constant. 
In turn, a change in the odds results in a change of the corresponding 
conditional outcome distribution, which puts higher probability mass on
larger age groups when changing $\B$ to $\B'$.
} \label{fig:streep}
\end{figure}
For evaluation, we use publicly available data of the actress Meryl Streep,
i.e., female sex and two images showing her at the age of 41 ($\B$, age group $[31, 46)$) 
and 67 ($\B'$, age group $[61, 117)$) to depict the predicted PDF and estimated log
odds-ratio in the SI-CS$_{\B}$-LS$_{\text{sex}}$ model (see Figure~\ref{fig:streep}).
The model yields the image-effect estimates $\eta(\B)=5.1$ and $\eta(\B')=10.1$,
while the effect of sex stays constant ($\beta_{\text{sex}}=0.3$). 
As expected $\eta(\B')>\eta(\B)$, indicating that $\B'$ is more likely to belong to
a higher age group than $\B$.
In particular, the difference between the two estimates yields a log odds-ratio 
$\eta(\B')-\eta(\B) = 5$,
which is interpretable as an $\exp(5)$-fold increase in the odds of belonging to
a higher age class compared to all classes below, when changing from $\B$ to $\B'$
and keeping sex constant. 

For a more systematic and empirical evaluation of the flexibility and interpretability 
of {\sc ontram}s, we fit seven models with the image data, the 10 simulated tabular 
predictors with known true effect sizes $\shiftparm$ and a combination of both 
(see Table~\ref{tab:models}).
The models differ in their flexibility due to different transformation functions and 
the parametrization of the loss. 
In Appendix~\ref{app:resqwk}, we compare the MCC model and the CI$_{\rx}$ {\sc ontram} 
to another ordinal classification model trained with a loss based on
Cohen's quadratic weighted kappa \citep[QWK,][]{de2018weighted}.

\begin{figure}[!ht]
\centering
\includegraphics[width=0.8\textwidth]{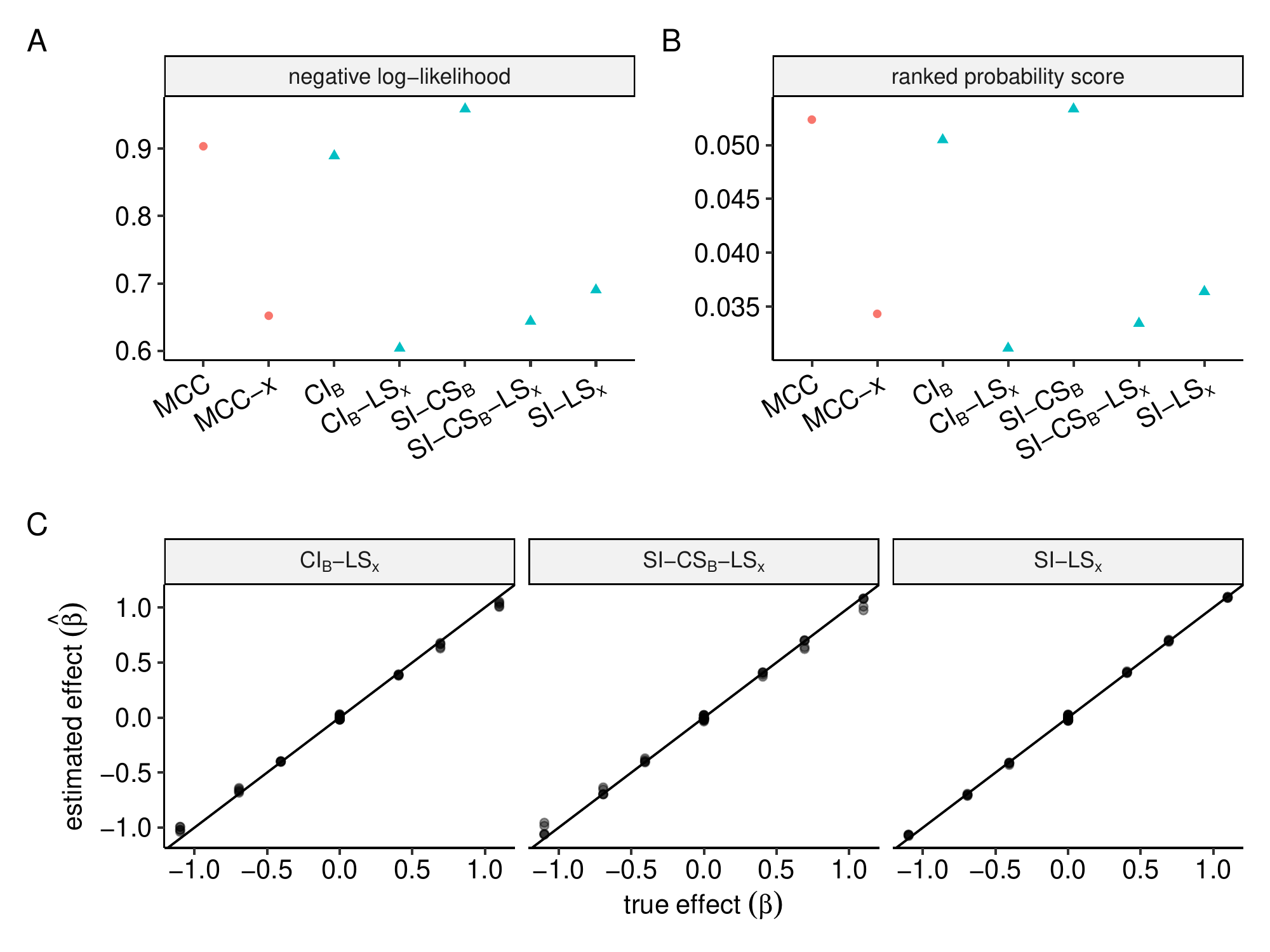}
\caption{
Test prediction performance for deep ensembles based on the UTKFace data. 
The figure summarizes the results for the models 
MCC, MCC-$\rx$, CI$_{\B}$, CI$_{\B}$-LS$_{\rx}$, SI-CS$_{\B}$, SI-CI$_{\B}$-LS$_{\rx}$,
SI-LS$_{\rx}$ (x-axes) in terms of negative log-likelihood (A)
and ranked probability score (B). Lower values in NLL and RPS indicate
improved model performance. Baseline models are highlighted in red,
{\sc ontram}s in blue.
C: True versus estimated predictor effects. The figure summarizes the true versus
estimated effects of the simulated tabular predictors of the UTKFace data set.
The effect estimates result from the linear shift terms, LS$_{\rx}$, in the models 
CI$_{\B}$-LS$_{\rx}$, SI-CS$_{\B}$-LS$_{\rx}$, SI-LS$_{\rx}$. 
In case of correct estimation, the parameters lie on the main diagonal.
} \label{fig:facesim}
\end{figure}

We first consider the most flexible models, MCC and CI$_{\B}$, which are based on the 
UTKFace image data and only differ in the parametrization of the loss function 
(see eq.~\ref{eq:cce} for the MCC and eq.~\ref{ontram_nll} for the {\sc ontram} loss).
As expected, the CI$_{\B}$ {\sc ontram} and 
MCC model achieve comparable prediction performances in terms of NLL and RPS (see Figure~\ref{fig:facesim}~A and B).
After including the simulated tabular predictors, the performance in both models increases notably
(see MCC-$\rx$ and CI$_{\B}$-LS$_{\rx}$ in Figure~\ref{fig:facesim}~A and B). 
In case of the MCC-$\rx$ model, the tabular predictors
are attached to the feature vector 
resulting from the convolutional part of the CNN, which allows interactions between
image and tabular predictors and therefore makes the model slightly more flexible 
than the CI$_{\B}$-LS$_{\rx}$.
However, in contrast to the CI$_{\B}$-LS$_{\rx}$, the MCC-$\rx$ allows no interpretation 
of the effect of the tabular predictors on the outcome.

Less flexible but more interpretable {\sc ontram}s are obtained by including the image
data as complex shift rather than as complex intercept term (SI-CS$_{\B}$). 
Although the SI-CS$_{\B}$ model is less flexible than the CI$_{\B}$ model, 
prediction performance is comparable (see Figure~\ref{fig:facesim}~A and B).
Again, adding the simulated tabular 
data as a linear shift term (SI-CS$_{\B}$-LS$_{\rx}$) results in improved prediction
performance.

Using a model with simulated tabular data only (SI-LS$_{\rx}$) yields a better 
performance than models that include image data only 
(see SI-LS$_{\rx}$ \emph{vs.} MCC, CI$_{\B}$, SI-CS$_{\B}$ in Figure~\ref{fig:facesim}~A and B). 
However, when comparing the models with image data and tabular predictors to the model with 
tabular predictors only, an increase in prediction performance is observed 
(see SI-LS$_{\rx}$ \emph{vs.} MCC-$\rx$, CI$_{\B}$-LS$_{\rx}$ and SI-CS$_{\B}$-LS$_{\rx}$).
This indicates that the images contain additional information for age prediction.

In practice, the {\sc ontram}s CI$_{\B}$-LS$_{\rx}$ and SI-CS$_{\B}$-LS$_{\rx}$
are most attractive because they provide interpretatable estimates for the effects
of the tabular predictors with an acceptably low decrease in prediction performance.

To assess whether effect estimates for the tabular predictors are reliable in models with and 
without additional image data, we compare the true effects $\beta$ to the
estimated effects $\hat{\beta}$ for the {\sc ontram}s with linear shift terms
(CI$_{\B}$-LS$_{\rx}$, SI-CS$_{\B}$-LS$_{\rx}$, SI-LS$_{\rx}$). 
As summarized in Figure~\ref{fig:facesim}~C, all models recover the correct
estimates.

\subsection{Wine Quality} \label{sec:reswine}
The experiments with the UTKFace data have shown that we get reliable and interpretable model
components when including simulated, mutually independent tabular predictors besides image data. 
In the following, we summarize a couple of experiments with the smaller wine data set containing
solely tabular predictors to demonstrate how we can estimate reliable linear and non-linear
effect estimates for potentially dependent tabular predictors.
In addition, we evaluate how the {\sc ontram} parametrization
of the loss (see eq.~\ref{ontram_nll}) yields a gain in training speed and
how this gain depends on the size of the training data.
Note that all those models can simply be extended to additionally 
include image data, e.g. by attaching a complex shift term CS$_{\B}$. 

\begin{figure}[!ht]
\centering
\includegraphics[width=0.8\textwidth]{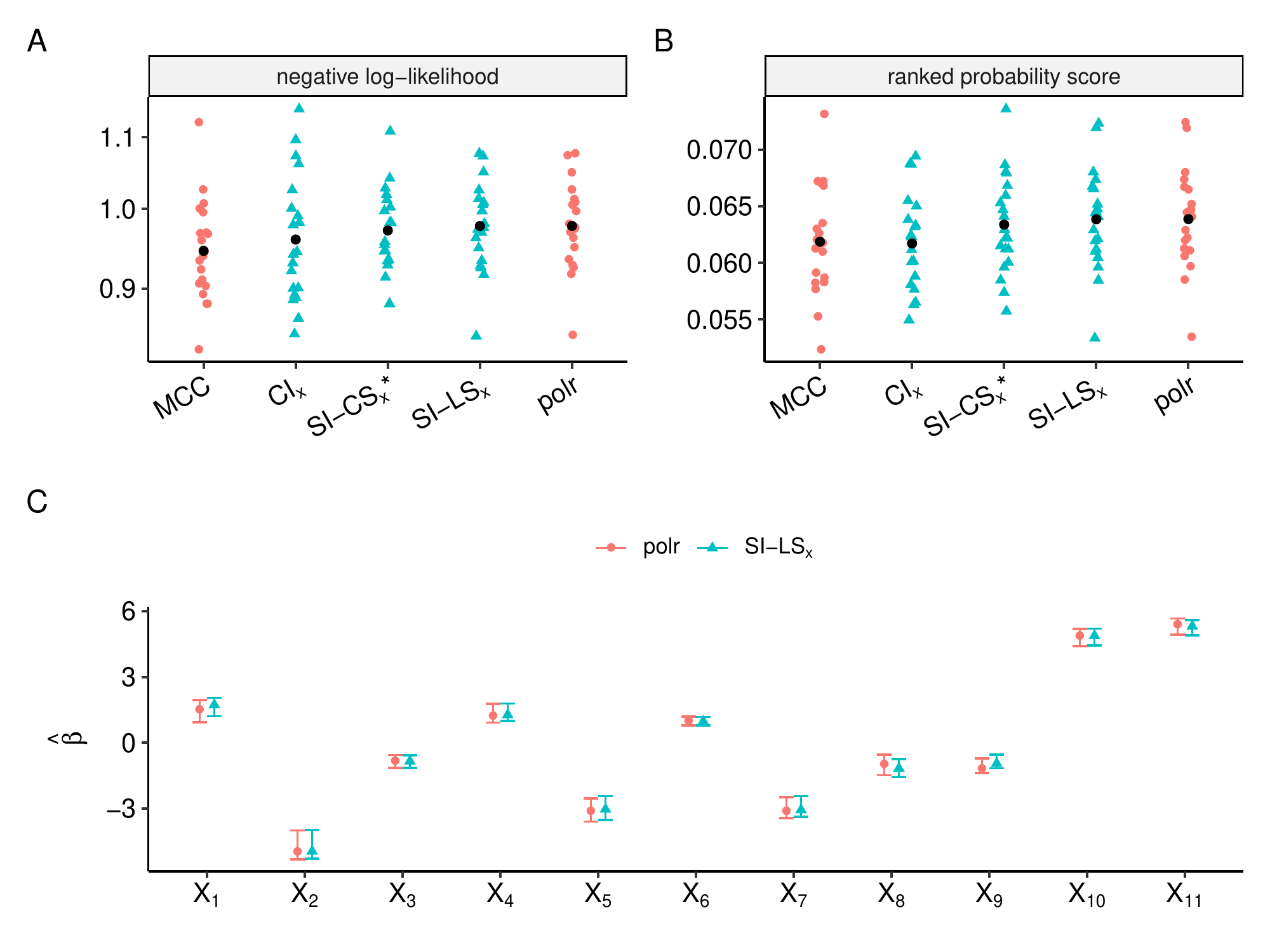}
\caption{
Results of the wine quality data based on the test sets from the cross validation setting.
Panels A and B summarize the prediction performance for the models MCC, CI$_{\rx}$, GAM,
SI-LS$_{\rx}$ and polr (x-axes) based on the wine quality data set in terms of negative
log likelihood (A) and ranked probability score (B). Lower values in NLL and RPS
indicate improved model performance. Results of {\sc ontram}s are indicated as blue triangles,
others as red dots. The black point gives the mean across the respective metric resulting
from the single CV folds. C: Effect estimates with 2.5th and 97.5th percentile for
polr and SI-LS$_{\rx}$ model over the 20 CV folds of the wine quality data set.
} \label{fig:predwine}
\end{figure}

The wine dataset is a benchmark data set for a proportional odds model that allows 
to interpret the fitted effect estimates as log odds-ratios (see Section~\ref{sec:interpretation}).
To illustrate the high flexibility of {\sc ontram}s and that we correctly estimate
linear, non-simulated tabular predictors, we fit a proportional odds model with linear 
effects via a SI-LS$_{\rx}$ model and compare the model to the same model 
using the \proglang{R} function \cmd{tram::Polr}.
As expected, Figure~\ref{fig:predwine} shows that both models yield the same prediction 
performance in terms of NLL (A) and RPS (B) and estimated predictor effects (C).

\begin{figure}[!ht]
\centering
\includegraphics[width=0.8\textwidth]{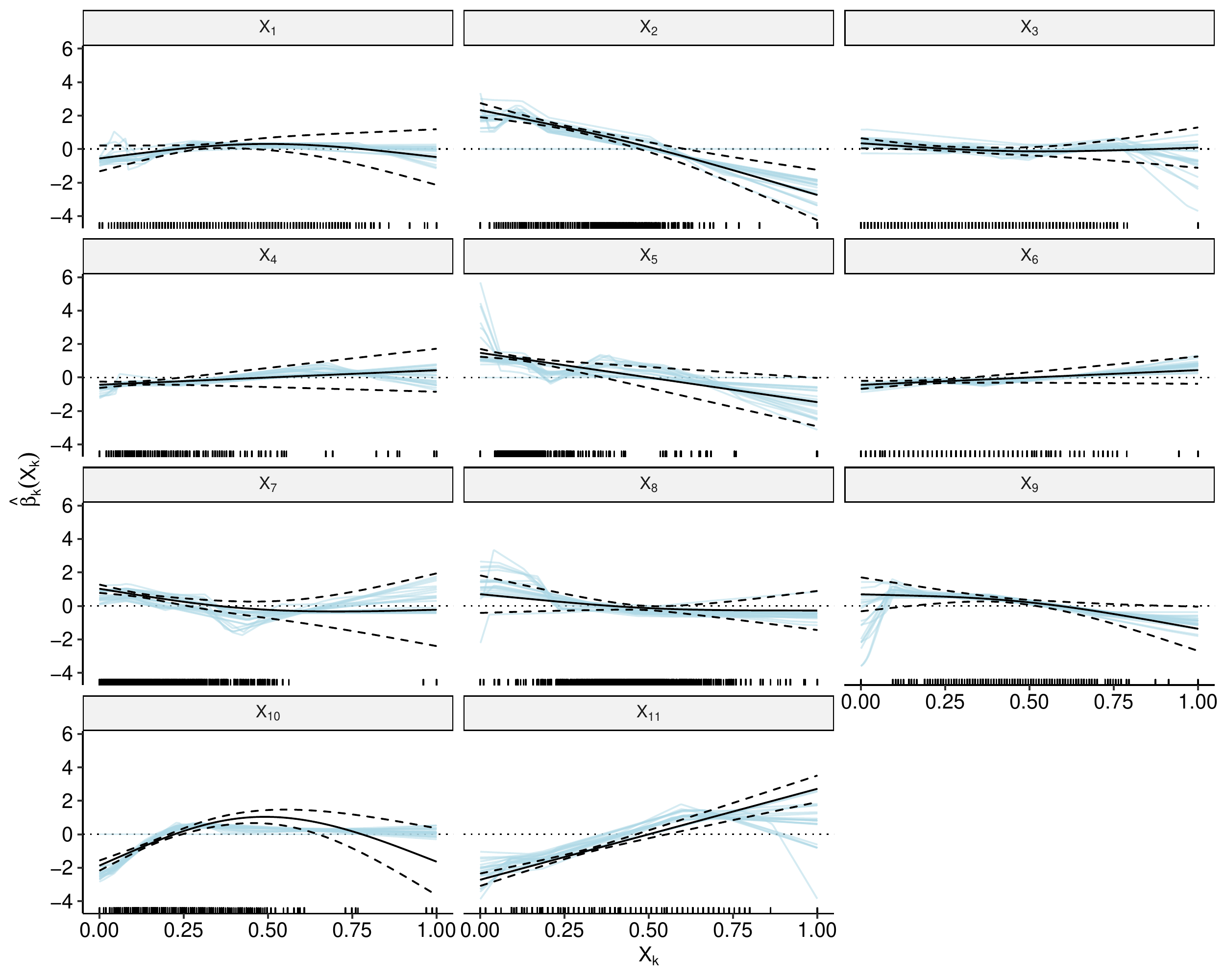}
\caption{
Estimated non-linear effects of 11 tabular predictors on the ordinal quality 
outcome in the wine data set as achieved by the {\sc ontram} GAM model.
The estimates log odds-ratio functions of an ensemble of 20 runs with different 
initial weights are shown in blue. 
The solid black line depicts the estimated log odds-ratio functions estimated
by the \texttt{mgcv::gam()} function in \proglang{R} together with a 95\% confidence interval
(dashed black lines). Rugs on the bottom of each plot indicate the observed values for
$X_k$, $k = 1, \dots, 11$, in the training data.
} \label{fig:gampred}
\end{figure}
GAMs (see Table~\ref{tab:models}, SI-CS$_{\rx}^*$ with 
$\h(\ry_k \given D)=\eparm_k - \sum_{j=1}^p \eshiftparm_j(x_j)$) 
add another layer of complexity to 
the model by allowing non-linear effects for each predictor. 
Because the individual NNs estimating the additive components $\beta_j(x_J)$ do not interact explicitly the
estimated log odds-ratio function retains the interpretability of a proportional
odds model. Figure~\ref{fig:gampred} depicts the estimates of an ensemble of {\sc ontram}
GAMs in comparison to a GAM from the \proglang{R}-package \pkg{mgcv}. Apart from
the constraint-enforced smoothness in \pkg{mgcv}'s GAM, both models agree in magnitude 
and shape of the estimated predictor effects. 
For instance, predictor $X_{10}$ (sulphate content) has a strong positive influence on 
the rating when increased from 0 to 0.25 (on the transformed scale),
in that the odds of the wine being rated higher increase by a factor of $7.4$, all
other predictors held constant
($\exp(\hat\eshiftparm_{10}(0.25) - \hat\eshiftparm_{10}(0)) \approx \exp(2) \approx 7.4$).
Afterwards the effect levels off and stays constant
for the {\sc ontram} GAM, due to regularization and few wines with higher sulphate levels
being present in the training data. The curve estimated by \pkg{mgcv} follows smoothness
constraints and instead drops with a large confidence interval, also covering 0.
GAMs are a special case of complex shift models, the latter of which allow for higher 
order interactions between the predictors.
Conceptually, {\sc ontram}s enable to further trade off interpretability and flexibility by 
modelling some predictor effects linearly while including others in a complex shift 
or intercept term.
If field knowledge suggests non-linearity of effects or interacting predictors, they
can be included as a complex shift or, if the proportionality assumption is violated,
in a complex intercept term.
From Figure~\ref{fig:gampred} we can see that most of the coefficients could be
safely modelled in a linear fashion, which is also evident from the minor loss
in predictive power when comparing the GAM against the linear shift {\sc ontram} 
(see Figure~\ref{fig:predwine}~A, GAM \emph{vs.} SI-LS$_{\rx}$).
\begin{figure}[!ht]
\centering
\includegraphics[width=0.8\textwidth]{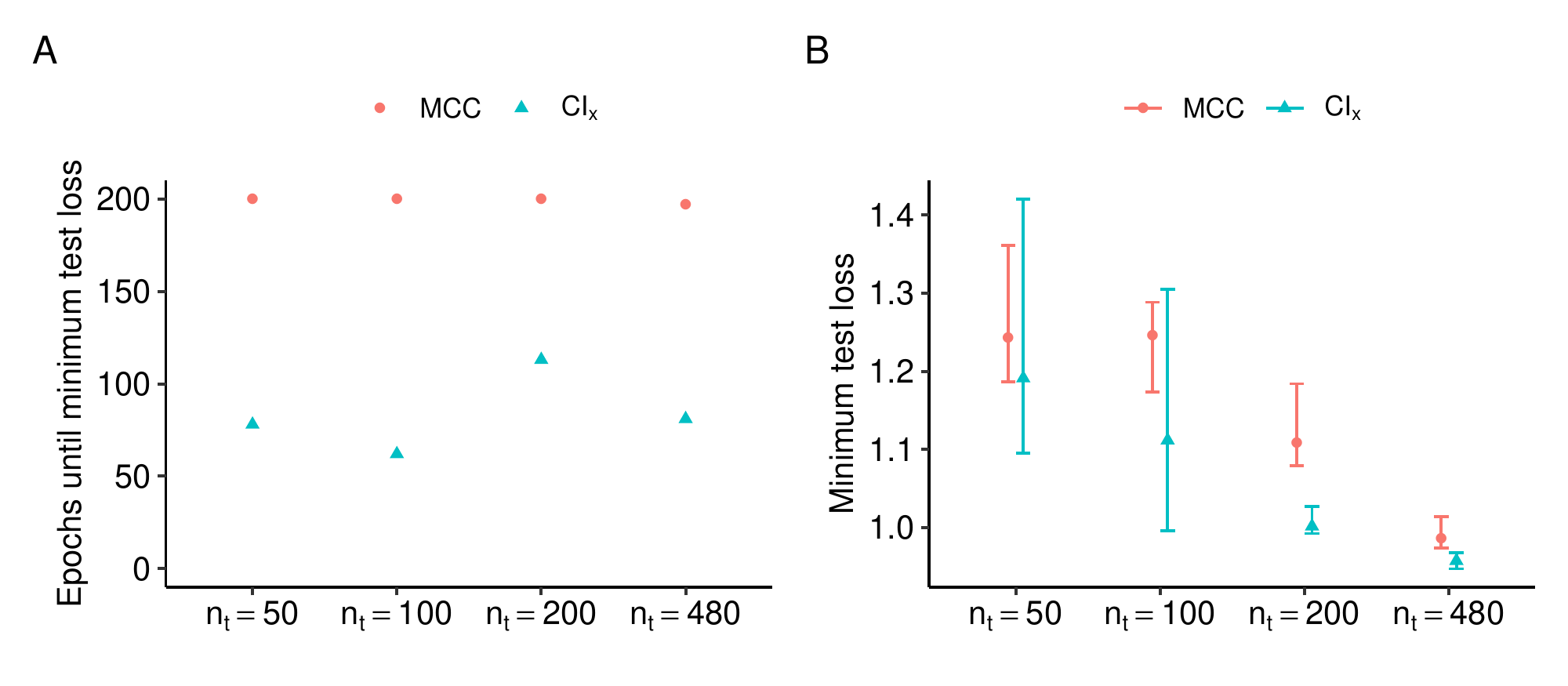}
\caption{
Epochs until minimum test loss for varying sizes of the training data using the
wine quality data set. The data are split into $n/n_t$, $n_t\in\{50, 100, 200, 480\}$
folds each of which serves as training data for a multi-class classification and a complex 
intercept {\sc ontram}.
The median test loss is computed for each scenario $n_t$ and each epoch.
Afterwards, the number of epochs until minimum median test loss
and the minimum median test loss are recorded.
Here the epochs until minimum test loss (A) and the minimum test loss (B)
are plotted against the 4 scenarios given by $n_t$.
} \label{fig:eff}
\end{figure}

To assess the effect of respecting the order of the ordinal outcome, we evaluate 
the most flexible CI$_{\rx}$ {\sc ontram} and the MCC model, which solely differ
in the parametrization of their loss. As in the UTKFace data, both models show the 
expected agreement in achieved prediction performance w.r.t. NLL and RPS
(see Figure~\ref{fig:predwine}~A and B). 
However, the CI$_{\rx}$ model learns much faster in terms of number of epochs until the 
minimum test loss is achieved, compared to the MCC model.
To further investigate this gain in learning speed, we split the wine quality
data into $n/n_t$, $n_t\in\{50, 100, 200, 480\}$ folds of size $n_t$ and fit a MCC model
and CI$_{\rx}$ {\sc ontram} to each fold.
The median test loss is computed for each scenario of size $n_t$.
The number of epochs needed to achieve minimal median test loss is summarized in
Figure~\ref{fig:eff}~A.
The training speed is consistently lower and therefore more efficient for the CI$_{\rx}$
{\sc ontram} than for the MCC model (Figure~\ref{fig:eff}~A).
The CI$_{\rx}$ {\sc ontram} yields a slightly better prediction performance (median test NLL)
for larger sample sizes. 
This can be explained by the fact that after 200 epochs the MCC model still has not
reached the minimum test loss (Figure~\ref{fig:eff}~B).
Note that the gain in training speed is only present if the outcome is truly ordered.
In Appendix \ref{app:learnspeed}, we show that the effect vanishes when the ordering of
the class labels is permuted.

\section{Discussion and Outlook} \label{sec:conclusion}

In this work we demonstrate how to unite the classical statistical approach to
ordinal regression with DL models to achieve interpretability of selected model components.
This allows us to estimate effects for the input data. 
In case of tabular predictors, we prove that the effects are correctly estimated, also in 
the presence of complex image data.
Moreover, we show that the most flexible {\sc ontram} trained with the reparametrized NLL 
achieves on-par performance with a MCC DL model using the cross-entropy loss.
This may first seem counter-intuitive because the cross-entropy loss ignores the outcome's 
order.
However, the {\sc ontram} NLL is a reparametrization of the cross-entropy loss and can, 
therefore, at most achieve the same performance.
The advantages of reparametrizing the NLL are (i) a natural scale for the additive and
hence interpretable decomposition of tabular and image effects, (ii) a valid probability 
distribution for the ordinal outcome and (iii) an increase in training speed.
In this context, interpretability is the main advantage over other state-of-the-art models 
because it is of crucial importance in sensitive applications as, for example, in medicine~\citep{rudin2018stop}.

If the focus lies mainly on classification of an ordinal outcome and less on interpretability 
and probabilistic predictions, the data analyst may be interested in optimizing a classification 
metric such as Cohen's kappa.
Indeed, Cohen's kappa directly considers the outcome's natural order and misclassifications 
further away from the observed class are penalized more strongly than misclassifications closer 
to the observed class.
However, this approach results in predictions different from those of regression models such 
as the MCC and CI$_{\B}$, which is further highlighted in Appendix~\ref{app:resqwk}.
In a regression model, on the other hand, the goal is rather to estimate a valid probability distribution which 
is achieved with proper loss functions such as the NLL.
These fundamental differences between ordinal classification and regression make a fair comparison 
nearly impossible, as we highlight in Appendix~\ref{app:resqwk}.

Further, we demonstrate how to select an {\sc ontram}, which possesses the appropriate amount
of flexibility and interpretability for a given application.
To achieve a higher degree of interpretability, flexibility has to be restricted, \eg by moving
from a complex intercept to a simple intercept, complex shift model.
However, we show that a restriction of flexibility can still yield adequate prediction performance 
which may even be similar to that of a more flexible model.
Interpretability of different model components is further showcased for simple
models including only tabular predictors and more complex models with tabular and image data.

The modular nature of {\sc ontram}s makes them highly versatile and applicable
to many other problems with ordinal outcome and complex input, such as text 
or speech data.
Instead of using a CNN for image data, a recurrent neural network can be used to define a more 
flexible complex intercept or a simpler, but more interpretable complex shift term as in a SI-CS$_\B$ 
{\sc ontram}.
Tabular predictors can then simply be added with linear shift or complex shift terms depending on 
the degree of interpretability the data analyst aims for.

This work shows the potential of deep transformation models for ordinal outcomes. 
The predictive power of deep transformation models on regression problems with 
continuous outcomes has already been demonstrated \citep{sick2020deep}.
However, the approach is easily extendable to the full range of existing interpretable
regression models, including models for count and survival outcomes.
The extension from ordinal data to count and survival data is hinted at by the parametrization
of the {\sc ontram} NLL, which can be viewed as an interval-censored log-likelihood
over the latent variable $\rZ$ for which the intervals are given by the conditional
cut points $\h(\ry_k \given D)$.
For count data these cut points are given by consecutive integers, i.e.,
$(0,1]$, $(1,2]$, and so on. In survival data the interval is given by
(commonly) right censored outcomes when a patient drops out of a study or experiences
a competing event. In case of right-censoring the interval is given by
$(t,+\infty)$ for a patient that drops out at time $t$.
All benefits in terms of interpretability and modularity will carry over to the deep 
transformation version of other probabilistic regression models by working with an 
appropriate likelihood and parametrizing the transformation function via (deep) neural networks.

\paragraph{Acknowledgements} 
We would like to thank Elvis Murina and Muriel Buri for insightful discussions and 
Malgorzata Roos for her feedback on the manuscript.
We thank all anonymous reviewers for their comments and suggestions, which helped
contextualize our proposed method among other state-of-the-art approaches.
The research of LH, LK and BS was supported by Novartis Research Foundation (FreeNovation~2019).
TH was supported by the Swiss National Science Foundation (SNF) under the project
``A Lego System for Transformation Inference'' (grant no. 200021\_184603).
OD was supported by the Federal Ministry of Education and Research of Germany (BMBF) 
in the project ``DeepDoubt'' (grant no. 01IS19083A).

\clearpage


\appendix
\renewcommand{\thesection}{\Alph{section}}
\counterwithin{figure}{section}
\renewcommand\thefigure{\thesection\arabic{figure}}
\counterwithin{table}{section}
\renewcommand\thetable{\thesection\arabic{table}}
\section{Neural Network Architectures} \label{app:NN}

We use NNs with different architectures to explore the various additive
decompositions of the transformation function described in Section~\ref{sec:ontram}.
Simple intercept and linear shift terms are always modeled with single layer NNs
without bias terms and a linear activation function
(see left panel in Figure~\ref{fig:ONTRAMarchitecture}).
For a fair comparison, we use the same NN architecture for all MCC
models and for controlling the complex intercept or complex shift term in an {\sc ontram}.
The only difference between the models is in the output layer. 
MCC models feature an output layer with $K$ units and a softmax
activation function. The output layer of complex intercept terms has $K-1$ output
units and a linear activation function. In case of complex shift terms, the output
layer consists of one unit and a linear activation function.

\begin{figure}[!ht]
\centering
\includegraphics[width=0.4\textwidth]{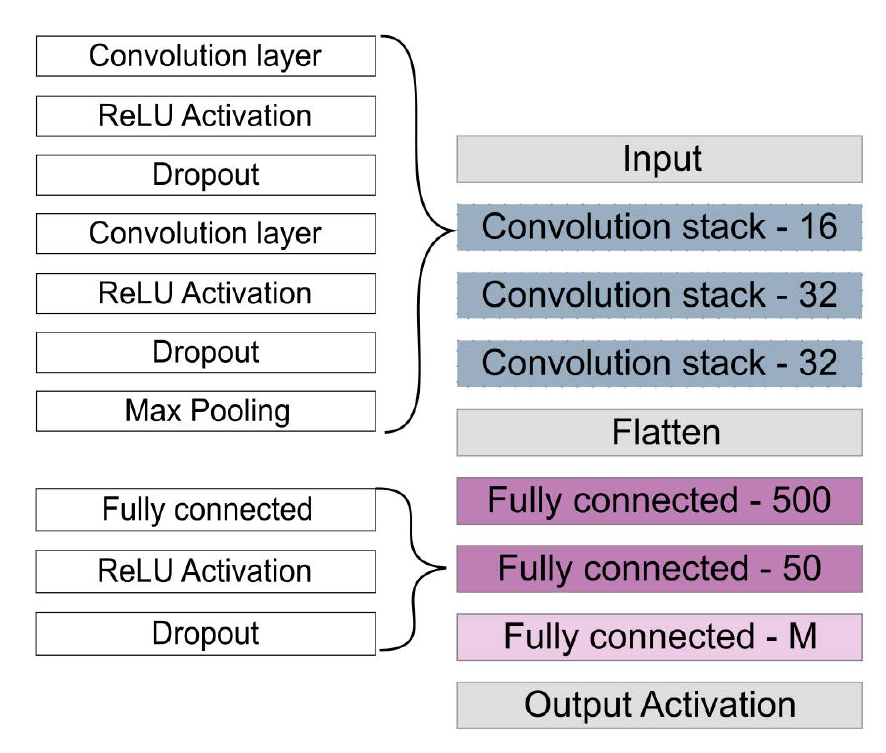}
\caption{
CNN architecture for controlling MCC models, complex intercept and
complex shift terms in case of image input data. 
The network architecture consists of stacks of convolutional blocks consisting of 
multiple repetitions of a convolutional, dropout and batch-normalization layer, 
followed by a fully connected part.
} \label{fig:cnn}
\end{figure}

\paragraph{UTKFace}
MCC, CI and CS terms are modeled with a CNN (see Figure~\ref{fig:cnn}).
The architecture used in this work is inspired by the VGG \citep{Simonyan2014vgg}.
The convolutional part consists of three blocks while
each block is comprised of two convolution and two dropout layers (with dropout rate 0.3).
The blocks complete with a max-pooling layer (window size $2\times2$ pixels, stride width 2).
In the first block we use 16 filters; the following two blocks contain 32.
Filter size is fixed to 3x3 pixels in every block.
The fully connected part features a 500- and a 50-unit fully connected layer.
The ReLU non-linearity is used as the activation function.

\paragraph{Wine quality}
MCC models, CI and CS terms are modeled using a densely connected neural
network with four layers having 16 units each. Between layers 1, 2, and 3, we
specify a dropout layer with dropout rate 0.3.
ReLU is used as the activation function. GAMs are fitted using the same
densely connected neural network for each predictor with two layers of 16 
units and followed by a layer of eight units with ReLU activation. 
All weights were regularized using both $L_1$ and $L_2$ penalties, with
$\lambda_1 = 1$ and $\lambda_2 = 5$.
Between each layer there is dropout with rate 0.2. 

\section{Learning Speed Under Permuted Class Labels} \label{app:learnspeed}

\begin{figure}[!ht]
\centering
\includegraphics[width=0.7\textwidth]{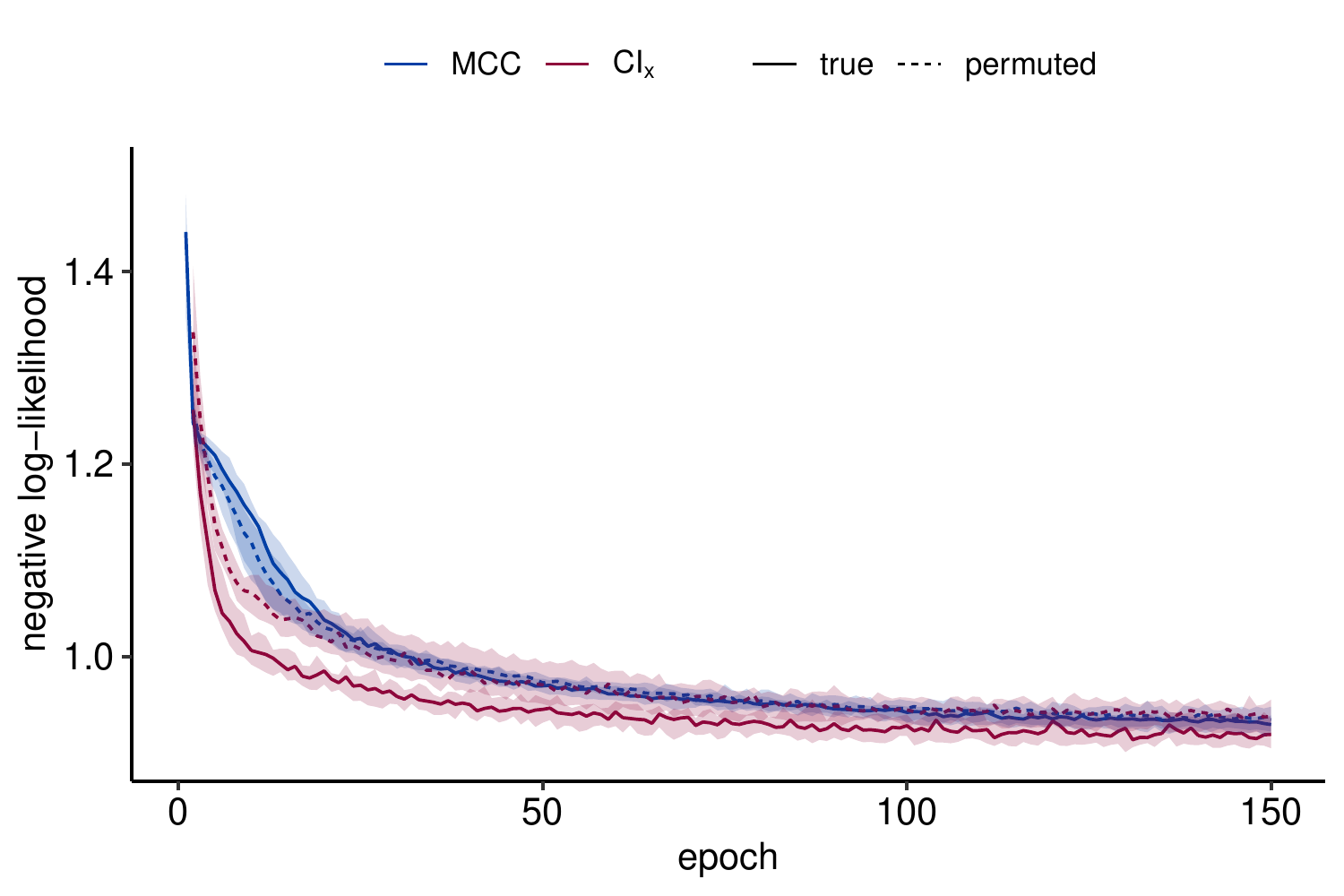}
\caption{
Test learning curves comparing a MCC against a complex intercept model (CI$_{\rx}$)
for the true and a permuted ordering of the classes in the wine quality data set.
Depicted are the median test losses (thick line) together with the empirical 20th
and 80th percentile (shaded regions) over 20 runs.
} \label{fig:permci}
\end{figure}

In contrast to MCC models with categorical cross-entropy loss, {\sc ontram}s' loss 
para\-metriza\-tion takes the ordering of the outcome into account (see eq.~\ref{ontram_nll}).
This leads to more efficient learning when the outcome is truly ordered.
This can be demonstrated by permuting the class label ordering of a truly ordinal outcome
and inspecting the learning curves of an {\sc ontram} and MCC model. When fitting
the model using the true class label ordering, the {\sc ontram} outperforms the MCC model in terms
of learning speed and arrives at virtually the same minimum test loss
(\emph{cf.} Figure~\ref{fig:permci}).
Permuting the class label order does not affect the learning curves of the MCC model.
However, the complex intercept {\sc ontram} performs on-par in the presence of wrongly
ordered categories in learning speed and median test loss compared to the MCC model.
In general, we observe a higher variability in the loss curves between different 
runs when using the {\sc ontram} loss parametrization.

\section{Interpretational Scales} \label{app:interpretation}
Classical regression textbooks, like \citet{tutz2011regression}, discuss
$\pZ \in \{\pSL, \pMEV, \pGumbel, \pN\}$, which are called cumulative logit,
minimum extreme value, maximum extreme value and probit models, respectively.
\citet{tutz2011regression} also derives the interpretational scales and stresses,
that the parameters resulting from a cumulative probit model are hard to interpret.
\begin{table}[ht!]
\centering
\caption{
Interpretational scales induced by $\pZ$ \citep{tutz2011regression}.
Same as Table~\ref{tab:interpr} in the main text, restated for convenience.
}\label{tab:interpretation}
\begin{tabular}{llll}
\toprule
\textbf{$F_Z$} & \textbf{$F_Z^{-1}$} & \textbf{Symbol} &
  \textbf{Interpretation of shift terms} \\ \midrule
Logistic & $\logit$ & $\pSL = \operatorname{expit}$ & log odds-ratio \\
Gompertz & cloglog & $\pMEV$ & log hazard-ratio \\
Gumbel & loglog & $\pGumbel$ & log hazard-ratio for $\rY_r = K + 1 - \rY$\\
Normal & probit & $\pN$ & not interpretable directly \\
\bottomrule
\end{tabular}
\end{table}

For $\pZ = \pSL := (1 + \exp(-z))^{-1}$ in a linear shift model $h(Y=y_k|x) = \eparm_k - \linpred $
\begin{align*}
\prob(\rY \leq \ry_k \given \rx) &= 
\pSL(\eparm_k - \linpred) \\
\Leftrightarrow  \logit(\prob(\rY \leq \ry_k \given \rx)) &= 
\eparm_k - \linpred \\
\Leftrightarrow  \log(\odds(\rY \leq \ry_k \given \rx)) &= 
\eparm_k - \linpred \\
\Leftrightarrow  \log(\odds(\rY \leq \ry_k \given \rx)) &= 
\log(\odds(\rY \leq \ry_k \given \rx=0)) - \linpred \\
\Leftrightarrow  -\log(\odds(\rY > \ry_k \given \rx)) &= 
-\log(\odds(\rY > \ry_k \given \rx=0)) - \linpred \\
\Leftrightarrow  \odds(\rY > \ry_k \given \rx) &= 
\odds(\rY > \ry_k \given \rx=0) \cdot \exp(\linpred), 
\end{align*}
which shows that the components of $\shiftparm$ are interpretable as log odds-ratios.
In the same way, $\eta(\B)$ in a complex shift model can be interpreted as a log odds-ratio function.
Table~\ref{tab:interpretation} summarizes the four most commonly used cumulative
ordinal regression models and the interpretational scales that different $\pZ$ induce.

For $\pZ = \pMEV = 1 - \exp(-\exp(z))$ we can interpret the components of $\shiftparm$ 
as log hazard ratios
\begin{align*}
1 - \pY(\ry \given \rx) &= \exp(-\exp(\h(\ry) + \linpred))
  = \exp(-\exp(\h(\ry))\exp(\linpred)) \nonumber \\
  &= \exp(-\exp(\h(\ry)))^{\exp(\linpred)} = (1 - \pY(\ry \given \rx = 0))^{\exp(\linpred)}.
\end{align*}
This interpretational scale is commonly used in survival analysis, 
Here, a log hazard ratio $\shiftparm$ for a shift from $0$ to $\rx$ is defined
in terms of the conditional survivor function
$S_{\rY}(y\given \rx) = 1 - \pY(\ry \given \rx)$ given by
$S_{\rY}(y\given \rx) = S_{\rY}(y\given \rx = 0)^{\exp(\linpred)}$.

\section{Ordinal Classification With Cohen's Kappa} \label{app:qwk}
Recently, DL models for ordinal classification were introduced using a transformed continuous 
version of the quadratic weighted kappa (QWK) as the loss function \citep{de2018weighted}. 
Cohen’s kappa is usually used to assesses inter-rater agreement
and corrects it for the expected number of agreements under independence
\begin{align}
\kappa = \frac{p_{\text{obs}} - p_{\text{exp}}}{1 - p_{\text{exp}}},
\end{align}
where $p_{\text{obs}}$ and $p_{\text{exp}}$ are the observed and expected proportion
of agreement under independence. 
They are computed from a confusion matrix, after dividing all entries
by the number of predicted instances, $p_{\text{obs}}$ is given by the sum of the
diagonal elements and $p_{\text{exp}}$ by the diagonal sum of the
product of the row and column marginals.
An additional weighting scheme enables to penalize misclassifications far
away from the observed class more strictly \citep{cohen1968weighted} than misclassifications 
close to the observed class
\begin{align} \label{eq:dqwk}
\kappa_w = \frac{\sum_{i,j}w_{ij}o_{ij} - \sum_{i,j}w_{ij}e_{ij}}{1 - \sum_{i,j}w_{ij}e_{ij}},
\end{align}
where $e$ denotes the expected and $o$ the observed number of agreements in row $i$ and
column $j$ of a confusion matrix. The weights $w_{ij} = \frac{\lvert i - j \rvert^q}{(K-1)^p}$
and the exponent $q$ control the amount of penalization for predictions farther
away from the observed class.
Because Cohen's kappa considers the ordinal nature of the outcome when weighted
accordingly, it has been proposed and used as a loss function for deep ordinal
classification problems \citep{de2018weighted,de2019deep,vargas2019deep,vargas2020cumulative}.
The QWK loss is defined as
\begin{align} \label{eq:cqwk}
\ell(p) = \log(1 - \kappa(p)),
\end{align}
where $p$ is the predicted probability distribution and denotes the quadratically ($q=2$)
weighted Cohen's kappa. Originally defined for count data, \citet{de2018weighted}
generalized Cohen's kappa to be computable from a probability density and so be used
as a loss function for MCC networks (when using softmax in the last layer). 
However, the QWK loss is an improper scoring rule and thus does not yield calibrated 
probabilistic predictions (cf.~Appendix~\ref{app:scoring}). 
As such, we demonstrate its limited use in estimating conditional distributions for 
ordinal outcomes in Appendix~\ref{app:resqwk}.
However, if one were solely interested in ordinal classification, the QWK loss is useful
to penalize misclassifications farther away from the observed class, where the
weighting scheme controls the amount of penalization. Ordinal regression and
classification are contrasted in more detail in Appendix~\ref{app:resqwk}.

To contrast ordinal regression models with pure classification models
we additionally use evaluation metrics for ordinal classifiers, such as the accuracy, 
continuous QWK
(the QWK loss transformed back to its original scale via $1-\exp(\ell(p))$, see eq.~\eqref{eq:cqwk}),
and discrete QWK (see eq.~\eqref{eq:dqwk}).
Neither accuracy nor the QWK metrics are proper scoring rules and both can lead to
misleading results when evaluating probabilistic models
(\emph{cf.} Appendices~\ref{app:scoring}~and~\ref{app:resqwk}).

\section{Scoring rules} \label{app:scoring}
Here we give a brief overview on scoring probabilistic predictions. In the end
we show that the QWK loss is an improper scoring rule.

Scoring rules are used to judge the prediction quality of a probabilistic model
and have their roots in information theory and weather forecasting 
\citep{gneiting2005weather}.
On average, a strictly proper scoring rule only takes its optimal value if the
predicted probability distribution corresponds to the
``data generating'' probability distribution. 
Assume that $p$ is the data generating probability density,
then a proper score $S$ fulfills for any probability density $q$
\begin{align}
\Ex_{Y \sim p}[S(p;Y)] \leq \Ex_{Y \sim p}[S(q;Y)].
\end{align}
A score is strictly proper when equality in the above holds iff $q = p$
\citep{gneiting2007strictly}.

\citet{brocker2007scoring} stress the importance of proper scoring rules for
honest distributional forecasts. The NLL is is a strictly proper scoring rule,
which is why we use it in the present work. Strict propriety of the
NLL can quickly be seen by plugging the NLL into the definition of a proper
scoring rule and reducing it to the Kullback-Leibler divergence
\begin{align}
\Ex_{Z \sim p}[-\log(q(Z))] &\geq \Ex_{Z \sim p}[-\log(p(Z))] \nonumber \\
\Leftrightarrow \; \Ex_{Z \sim p}[-\log(q(Z))] - \Ex_{Z \sim p}[-\log(p(Z))] &\geq 0 \\
\Leftrightarrow \; \Ex_{Z \sim p}[\log(p(Z)/q(Z))] = \KL(p \rVert q) &\geq 0 \nonumber
\end{align}
which is indeed larger than 0 and equality holds iff $p = q$ \citep{kullback1951information}.
Indeed, any affine transformation of the NLL is a strictly proper scoring rule,
which follows directly from linearity of the expected value. An example would be a weighted
NLL in which the weights are chosen inversely proportional to the frequency of
the different outcome classes. 

As noted in the main text, only the probability assigned to the true class enters
the likelihood function. If a score is evaluated only at the observed outcome the
score is called local. To use a non-local proper scoring rule alongside
the local NLL that takes into account
the whole predicted conditional distribution of an ordinal outcome we 
use the ranked probability score (RPS). The RPS is defined as
\begin{align}
\RPS(p;y) = \frac{1}{K-1} \sum_{k=1}^K \left( \sum_{j=1}^k p_j - \sum_{j=1}^k e_j \right)^2,
\end{align}
where $K$ denotes the number of classes, $p_j$ and $e_j$ are the predicted and
actual probability of class $j$, respectively.
Note, that in most cases $e_j$ is given by the $j$th entry of the one-hot encoded
outcome. By summing over all $K$ classes the RPS incorporates the whole predicted
probability distribution and thus is a non-local scoring rule.
The RPS is a proper scoring rule, which can be shown by reducing it to the sum of
individual Brier scores at the classes $k = 1, \dots, K$. The Brier score is a
proper score for binary outcomes and given by
\begin{align}
\BS(p;y) = (y - p)^2.
\end{align}
By introducing another predicted probability $q \in (0,1)$, the expectation w.r.t.
$p$ can be written as
\begin{align}
\Ex_{Y \sim p}[(Y-p)^2] = \Ex_{Y \sim p}[(Y - q + q - p)^2] =
    \Ex_{Y \sim p}[(Y - q)^2] - (p - q)^2.
\end{align}
where $(p-q)^2$ will always be positive and hence
\begin{align}
\Ex_{Y \sim p}[(Y-p)^2] \leq \Ex_{Y \sim p}[(Y - q)^2]
\end{align}
which shows propriety of the Brier score. The last step is now to sum up the $K$
individual Brier scores to arrive at the RPS. Because each Brier score is proper,
the sum will also be minimal for the data generating distribution of the ordinal
outcome. However, the RPS is not strictly proper because individual probabilities
can be switched without changing the overall sum.

Some scoring rules possess a less desirable property in that they encourage overly
confident predictions by assigning a too high probability to the predicted outcome
value and thus fail to capture the actual uncertainty. Such a score is called
improper and examples include the linear score and mean square error
\citep{brocker2007scoring}.
Although the accuracy is a commonly used metric, its use is neither recommended for
assessing probabilistic predictions nor classification performance. This is because it is not a
scoring rule and formulating it as one requires additional assumptions, which may
very well yield an improper scoring rule. 
Also the QWK loss (see Appendix~\ref{app:qwk}),
\begin{align}
\ell(p) = \log(1 - \kappa(p)),
\end{align}
as proposed by \citet{de2018weighted} is improper.
This can be seen by taking the data generating density of an ordinal random variable
$Y\in\{y_1<y_2<y_3\}$ to be
\begin{align*}
p(y) =
  \begin{cases}
    0.3 & \mbox{if } y = y_1, \\
    0.4 & \mbox{if } y = y_2, \\
    0.3 & \mbox{if } y = y_3;
  \end{cases}
\end{align*}
for which the expected score will be
\begin{align}
  \Ex_p(\log(1 - \kappa(p))) = \sum_{k = 1}^K p(y_k) \log(1 - \kappa(p))
  = -0.693.
\end{align}
However, by forecasting a density $q$ that puts more mass on $p$'s mode, $Y = y_2$,
\begin{align}
q(y) =
  \begin{cases}
    0.1 & \mbox{if } y = y_1, \\
    0.8 & \mbox{if } y = y_2, \\
    0.1 & \mbox{if } y = y_3;
  \end{cases}
\end{align}
we can achieve a much lower (better) score
\begin{align}
  \Ex_{Y \sim p}(\log(1 - \kappa(q))) = \sum_{k = 1}^K
    p(y_k) \log(1 - \kappa(q))
  = -1.386
\end{align}
which proves impropriety of the QWK loss by counterexample.
The same argument holds for $K = 2$ and $K > 3$, as well as different weighting
schemes (no weighting, linear weights, higher order weights).

\section{Contrasting Ordinal Regression and Classification} \label{app:resqwk}
Here, we assess the benefit of using a proper versus an improper score for training
and contrast ordinal regression (predicting a distribution) versus classification
(predicting a single class or level). 
To this end, we compare {\sc ontram}s against a recently developed ordinal classifier, 
which uses the quadratic weighted kappa (QWK) loss \citep{de2018weighted} 
(see Appendix~\ref{app:qwk}) and the adapted $K$-rank approach DOEL$_2$ \citep{xie2019deep}.
While the discussion so far was focusing on ordinal regression models, predicting
a faithful conditional probability distribution, the developed (QWK-based) methods
primarily aim for predicting the correct class and penalize misclassifications 
by their distance to the correct class. 
The QWK loss is an improper scoring rule and does not make an honest probabilistic 
prediction but encourages overly confident predictions 
(as we show in Appendix~\ref{app:scoring}).
In contrast to QWK, DOEL$_2$ is a pure ordinal classifier and thus predicts
only a single class. Consequently, one cannot evaluate DOEL$_2$
using any proper score.
We emphasize again the difference between taking a regression versus a classification
approach to problems with an ordinal outcome. For that we use different
performance measures to compare the MCC model for probabilistic multi-class
classification, with the QWK model for ordinal classification and the CI {\sc ontram}s  (CI$_{\rx}$ 
for wine quality and CI$_\B$ for UTKFace in Table~\ref{tab:qwk-metrics}).
In (ordinal) classification tasks it is common to report the test accuracy and
quadratic weighted kappa, although they are improper scoring rules 
(Appendix~\ref{app:scoring}).

While it is unfair by design to compare ordinal regression models trained with NLL
loss, and ordinal classification models trained with a QWK loss, in terms of NLL
or QWK-based evaluation metrics, such comparisons are commonly seen. 
The QWK model shows a strong performance deficit
in terms of NLL and RPS compared to the MCC and CI models for all three data sets,
because the QWK loss is improper (see Table~\ref{tab:qwk-metrics}).
Conversely, we observe a better continuous quadratic weighted kappa for the QWK
model. These results are expected due to the different optimization criteria of
the models but can be misleading at first sight. The discrete QWK metric shows
almost on-par performance of all three models for the wine quality, while for the
UTKFace data the QWK model performs worst.
For DOEL$_2$ (in fact any ordinal classifier, which does not predict a full outcome
distribution, such as SVMs), NLL, RPS and continuous QWK cannot be computed.
In turn, one can only compare {\sc ontram}s against these ordinal classifiers in
terms of discrete QWK and accuracy.

\begin{table}[!ht]
\caption{
Test performance of the MCC model for probabilistic multi-class classification,
the QWK model for ordinal classification and the CI (complex intercept) {\sc ontram}
for ordinal regression in terms of proper scores (NLL, RPS), and for ordinal classification
in terms of classification metrics (continuous and discrete QWK, and
accuracy (Acc)) for the wine quality and UTKFace data sets.
The continuous QWK is obtained by back-transforming the QWK loss to its original 
scale via $1-\exp(\ell(p))$. All models were trained using the same NN architecture,
differing only in the last layer for an appropriate input to the loss function.
}\label{tab:qwk-metrics}
\resizebox{\textwidth}{!}{
\begin{tabular}{l|l|rr|rrr}
\toprule
\textbf{Data set} & \textbf{Model} & \textbf{NLL} & \textbf{RPS} & \textbf{cont. QWK} & \textbf{discr. QWK} & 
\textbf{Acc} \\
 \midrule
 & CI$_\B$ & \textbf{0.889} & \textbf{0.050} & 0.772 & \textbf{0.889} & \bf 0.638\\
 UTKFace & MCC & \textbf{0.903} & \textbf{0.052} & 0.769 & 0.879 & \textbf{0.633}\\
 & QWK \citep{de2018weighted} & 10.582 & 0.101 & \textbf{0.793} & 0.794 & 0.514\\
\midrule
 & CI$_{\rx}$ & \textbf{0.964} & \textbf{0.062} & 0.419 & 0.543 & 0.597\\
 Wine Quality & MCC & \textbf{0.948} & \textbf{0.062} & 0.375 & 0.532 & 0.605\\
 & QWK \citep{de2018weighted} & 5.713 & 0.094 & \textbf{0.566} & \textbf{0.575} & 0.564\\
 & DOEL$_2$ \citep{xie2019deep} & - & - & - & \bf 0.576 & \bf 0.615\\
\bottomrule
\end{tabular}
}
\end{table}

Albeit not an ordinal evaluation metric we compare the models in terms of classification
accuracy, as this is commonly done in benchmarking (ordinal) classification models.
The accuracy is misleading for various reasons. It is discontinuous at an implicitly
used threshold probability and yields misleading results if classes are strongly
unbalanced.
In the comparison in Table~\ref{tab:qwk-metrics} the QWK model shows the worst
and the MCC model best accuracy closely followed by the CI {\sc ontram} for all three
data sets. 
Both the theoretical considerations and the inconclusive results in Table~\ref{tab:qwk-metrics}
indicate that it is advantageous to refrain from
using classification accuracy to evaluate ordinal regression or classification models.

Before closing this section, we want to stress the advantage of using proper
scoring rules for assessing the prediction performance of an ordinal regression
model that focuses on predicting a whole conditional distribution, embracing
the uncertainty in the prediction. Also the MCC model is a probabilistic model
and can therefore be assessed via proper scoring rules. Solely the QWK model
is focusing on ordinal classification and therefore care should be taken when
evaluating it with both proper and improper scoring rules and comparing it to
ordinal regression models.

\section{Confusion Matrices and Accuracy UTKFace} \label{app:confmats}

\begin{figure}[!ht]
\centering
\includegraphics[width=0.8\textwidth]{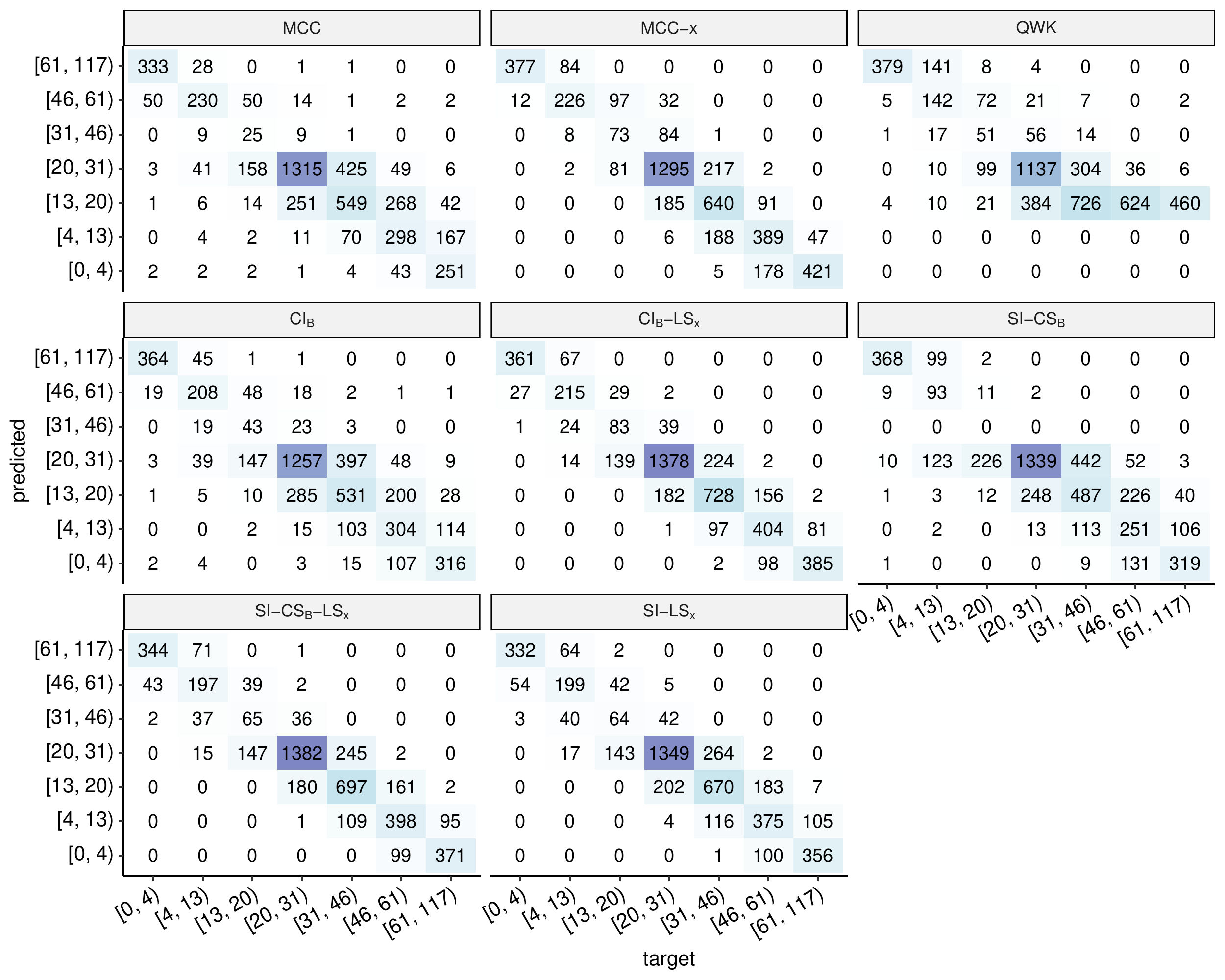}
\caption{
Confusion matrices for UTKFace data for all models mentioned in Table~\ref{tab:models}.
The figure summarizes the true ($x$-axis) 
vs. the predicted ($y$-axis) classes for the fitted models. 
The predicted class is the class with the highest predicted probability.
} \label{fig:confface}
\end{figure}

\begin{figure}[!ht]
\centering
\includegraphics[width=0.6\textwidth]{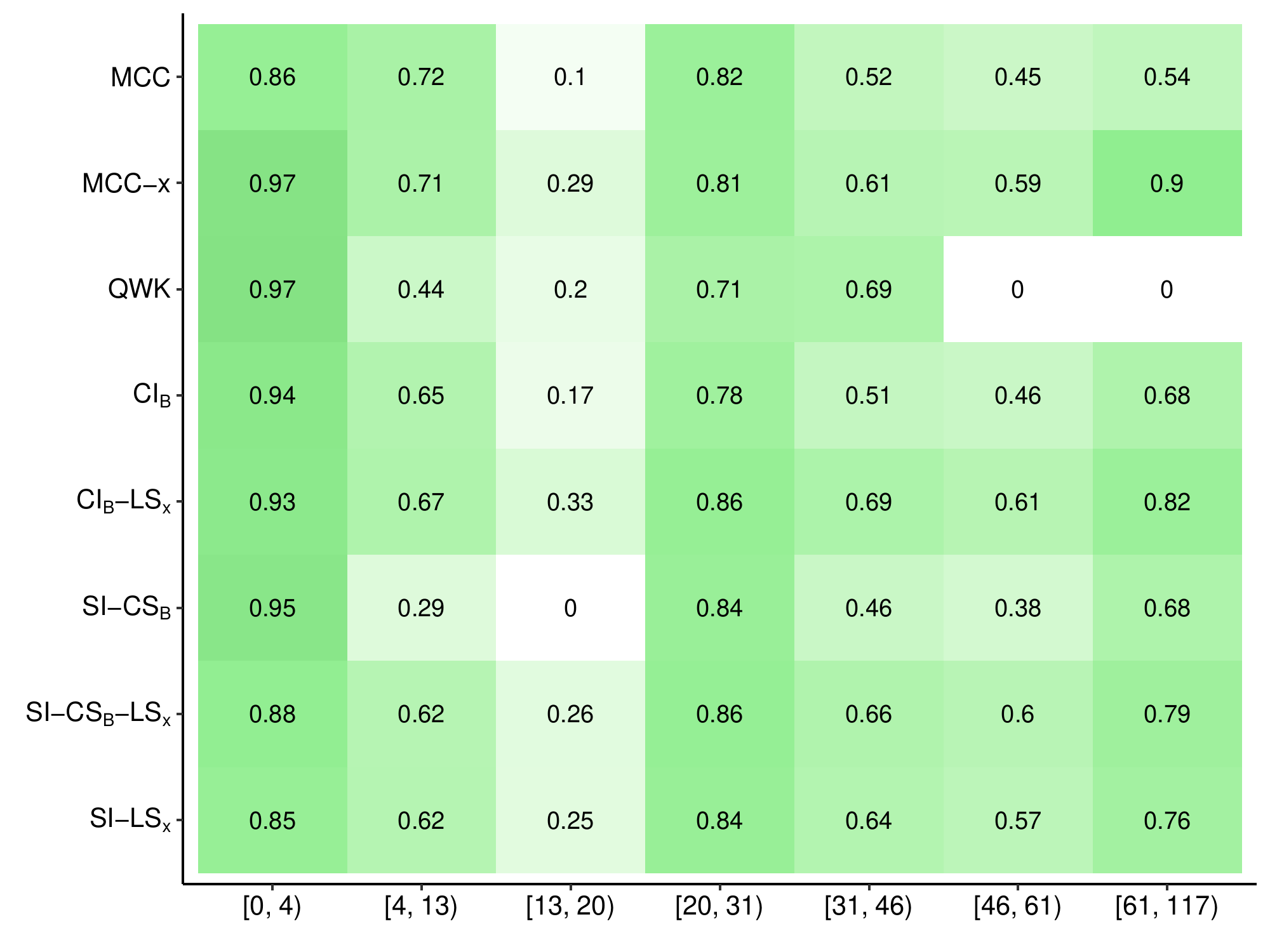}
\caption{
Class-wise accuracies of all models discussed in the main text for the UTKFace data.
} \label{fig:classaccface}
\end{figure}

UTKFace shows a moderately imbalanced marginal distribution 
of the class levels (7.99\% baby, 6.41\% child, 4.98\% teenager, 34.04\% young adult, 
22.92\% adult, 13.57\% middle aged, 10.1\% senior).
\begin{table}[!ht]
\centering
\caption{
Outcome distribution in the UTKFace test data set stratified by sex.
} \label{tab:UTKFace}
\begin{tabular}{@{}llllllll@{}}
\toprule
             & \multicolumn{7}{c}{\textbf{Age group}}  \\ \midrule
\textbf{Sex} & $[0, 4)$  & $[4, 13)$  & $[13, 20)$  & $[20, 31)$  & $[31, 46)$  & $[46, 61)$  & $[61, 117)$   \\
male         & 194 & 135 & 112 & 666 & 640 & 480 & 269 \\
female       & 195 & 185 & 139 & 936 & 411 & 180 & 199 \\ \bottomrule
\end{tabular}
\end{table}
Figure~\ref{fig:confface} shows the confusion matrices for all fitted models. 
Class-wise accuracies are depicted in Figure~\ref{fig:classaccface}. 

The MCC and QWK models as well as the {\sc ontram}s yield similar 
confusion matrices and class-wise accuracies across all models.


\clearpage

\vskip 0.2in
\bibliographystyle{plainnat}
\bibliography{mybibfile}

\end{document}